\documentclass[final,3p,times,sort&compress]{elsarticle}

\usepackage{graphicx}
\usepackage{algorithm2e}
\usepackage{amsmath}
\usepackage{amsthm}
\usepackage{hyperref}
\usepackage{todonotes}
\usepackage{subfigure}

\usepackage{amssymb}
\usepackage[english]{babel}
\usepackage{epstopdf}
\usepackage{graphicx}

\usepackage{verbatim}
\usepackage{todonotes}
\usepackage{comment}
\usepackage{adjustbox}
\usepackage{balance}
\usepackage{comment}

\usepackage{ulem}

\journal{Artificial Intelligence}

\newtheorem{mydef}{Definition}

\newcommand{\snrurl}{\url{https://drive.google.com/open?id=0BxZBpceSOYIRWkphMXZ5QTV2ajQ}}
\newcommand{\crdurl}{\url{https://drive.google.com/open?id=0BxZBpceSOYIRZlFBYkdQWURhbms}}

\newcommand{\dlol}{$DLOL$}

\newcommand{\Tau}{\mathfrak{I}}

\newcommand{\mids}{\,{\mid}\,}
\newcommand{\ins}{\,{\in}\,}

\newcommand{\ges}{\,{\ge}\,}
\newcommand{\les}{\,{\le}\,}

\newcommand{\eqs}{\,{=}\,}

\newcommand{\modelss}{\,{\models}\,}

\newcommand{\cups}{\,{\cup}\,}
\newcommand{\caps}{\,{\cap}\,}
\newcommand{\subseteqs}{\,{\subseteq}\,}
\newcommand{\sqsubseteqs}{\,{\sqsubseteq}\,}
\newcommand{\hs}{\hspace{5pt}}
\newcommand{\specialspace}{\hspace{0.3em}}
\newcommand{\specialpar}[1]{%
  \par\hangindent=1.5em \hangafter=1
  {#1: }\ignorespaces}
\newtheorem{theorem}{Theorem}

\def\bD{\mathbf{D}}

\def\cI{{\cal I}}

\def\cT{{\cal T}}

\begin{document}

\begin{frontmatter}



\title{Formal Ontology Learning from English IS-A Sentences}


\author[a]{Sourish Dasgupta}
\author[b]{Ankur Padia}
\author[c]{Gaurav Maheshwari}
\author[d]{Priyansh Trivedi}
\author[e]{Jens Lehmann}

\address[a]{sourish@rygbee.com, Rygbee, India}
\address[b]{padiaankur@gmail.com, University of Maryland, Baltimore County, USA}
\address[c]{gaurav.rygbee@gmail.com, DA-IICT, India}
\address[d]{priyansh.rygbee@gmail.com, DA-IICT, India}
\address[e]{jens.lehmann@cs.uni-bonn.de , University of Bonn, Germany}


\begin{abstract}
Ontology learning (OL) is the process of automatically generating an ontological knowledge base from a plain text document. In this paper, we propose a new ontology learning approach and tool, called DLOL, which generates a knowledge base in the description logic (DL) $\mathcal{SHOQ}(\mathbf{D})$ from a collection of factual non-negative IS-A sentences in English. We provide extensive experimental results on the accuracy of DLOL, giving experimental comparisons to three state-of-the-art existing OL tools, namely Text2Onto, FRED, and LExO. Here, we use the standard OL accuracy measure, called \textit{lexical accuracy}, and a novel OL accuracy measure, called \textit{instance-based inference model}. In our experimental results, DLOL turns out to be about 21\% and 46\%, respectively, better than the best of the other three approaches.	
\end{abstract}

\begin{keyword}
Knowledge modeling \sep Language parsing \& understanding \sep Ontology design \sep Knowledge Representations




\end{keyword}

\end{frontmatter}

\section{Introduction}
Comprehensive and consistent domain-specific ontologies are useful in building applications that can perform complex tasks such as question answering \cite{2moldovan2007lymba}, machine translation \cite{3DBLP:journals/expert/MaedcheS01}, bio-medical knowledge mining \cite{abeyruwan2013prontolearn}, and other Semantic Web applications. However, manually creating ontologies that may comprise of a large number of concepts and relations is a highly labor intensive and expensive task. With the scale of textual information present on the Internet, it is imperative to devise automated systems that can perform this task. This process, in which a system can generate ontologies out of unstructured Natural Language (NL) text, is called \textit{Ontology Learning} (\textit{OL}).

Depending on the degree of formal expressivity, existing OL techniques can be classified into two broad approaches: (i) \textit{lightweight OL} and (ii) \textit{formal OL}\footnote{Terms borrowed from the survey paper \cite{4wong2012ontology}.}. Lightweight OL generates \textit{taxonomies} of concepts that are automatically classified in terms of their mutual hyponymic (i.e. “\textit{IS-A}”) relations. On the other hand, formal OL automatically translates NL sentences into formal expressions (eg, Predicate Logic, Logic Form Translation (LFT) \cite{81moldovan2001logic}, Discourse Representation Structures (DRS) \cite{75kamp1981theory}, Description Logics \cite{baader2005description}, etc.), and then, constructs a generalized ontology using some kind of formal reasoner (eg. theorem provers, semantic-tableau based reasoners, etc.). Translation into formal expression is usually based on the \textit{principle of compositional semantics}\footnote{\url{http://plato.stanford.edu/entries/compositionality/}}, which states that meaning of a larger linguistic construct is a function of the meaning of its constituent elements and set of combinators (i.e. rules or functors) that governs how the lower-level meanings are combined. However, the translation process is not straight-forward. This is because there are several commonly observed linguistic nuances that violate the compositionality principle (we call them non-trivial sentences; see section \ref{is-a type} for further elucidation). For non-trivial sentences, we need to meticulously categorize a set special lexico-syntactic patterns and then introduce additional special translation combinators (along with the generic ones). 

In this paper, we limit the scope to factual IS-A sentences in English in order to develop a sufficiently robust theoretical foundation for formal OL on text. We, thereby, exclude figurative and allegorical sentences, such as ``\textit{John is a hard nut}". We show that OL on factual IS-A sentences is non-trivial in nature. Some of the linguistic non-trivialities that are commonly observed are: (i) lexical variations of quantifiers and ``\textit{is a}" lexemes, (ii) lexico-syntactic variations of IS-A patterns such as some of the Hearst patterns \cite{47hearst1992automatic}, (iii) tense variations of ``\textit{is a}" lexemes such as ``\textit{was}", "\textit{had become}", (iv) modal variations of ``\textit{is a}" lexemes such as ``\textit{may be}", ``\textit{could be}", and (v) comparative and superlative constructs of IS-A sentences. We demonstrate that OL on IS-A sentences provides the axiomatic foundation for formally representing factual non IS-A sentences\footnote{An example of factual non-ISA sentence is: ``\textit{The sun rises in the east.}"} as well, and therefore requires special study. To support this argument, we created a community IS-A test dataset and ran contemporary state-of-the-art general-purpose OL tools, but with unsatisfactory results with respect to standard accuracy measures. We propose a novel Description Logic (DL) \cite{baader2005description} based OL framework, called \textit{DLOL}, that automatically translates NL sentences into DL expressions, thereby generating both definitional and axiomatic ontology in OWL\footnote{\url{https://www.w3.org/TR/owl-features/}} format.

DLOL has a two-phase pipeline. In the first phase, it uses an intermediate canonical representation form, called \textit{Normalized Sentence Structure} (\textit{NSS}). One of the primary objectives of NSS is to normalize English sentences, having syntactic and lexical variations, into canonical NSS instances.  In the second phase, a \textit{DL Translation} (\textit{DLT}) algorithm converts NSS instances into their equivalent DL expressions, by mapping linguistic constructs into generic DL rules and axioms. Normalization using NSS helps the DLT algorithm to be less sensitive to structural variations, thereby increasing the accuracy of DL representation. The core contributions of this paper are as follows:

\begin{itemize}
	\item We show that for most factual IS-A type sentences the DL fragment $\mathcal{SHOQ}(\mathbf{D})$ is a sufficient representation language. 
	\item A novel paraphrase resilient sentence normalization structure (and a corresponding template-fitting algorithm), called NSS, is proposed. It is less sensitive to lexico-structural variations and also supports complex sentences thus serving as a robust intermediary formal sentence representation.
	\item An NSS instance to DL expression translation algorithm, called DLT, is proposed.
	\item An evaluation of DLOL in terms of: (i) syntactic robustness of NSS using WCL v1.1 (wiki\_good.txt) test corpus \cite{14navigli2010annotated}
 and domain specific ontologies from YAHOO! BOSS by \cite{49kozareva2010semi}, and (ii) semantic accuracy of DLT algorithm on the community released dataset and using gold-standard engineered ontology built on same WCL v1.1 corpus. We compared DLOL with three state-of-the-art existing OL tools, namely Text2Onto (lightweight OL) \cite{52cimiano2005text2onto}, FRED (formal OL) \cite{13presutti2012knowledge,Gangemi2016}, and LExO (formal OL) \cite{volker2007acquisition}, using a standard OL evaluation measure, called Lexical Accuracy (LA) \cite{85dellschaft2006perform}. We also introduce a novel benchmark measure, called \textit{Instance-based Inference Measure} (\textit{IIM}). We observed that DLOL had an average recall improvement of about 21\% and 46\% (in terms of LA and IIM, respectively), when compared to the next best performing tool. 	
	
\end{itemize}
The rest of this paper is organized as follows: (2) \textit{Preliminaries}, where we introduce basic concepts about IS-A sentences and the DL $\mathcal{SHOQ}(\mathbf{D})$, (3) \textit{Problem Statement}, where we provide the motivation for formal OL and define it, (4) \textit{Approach Overview}, where we outline the motivation and architectural pipeline of DLOL, (5) \textit{NL Sentence to NSS Instance}, where we provide formal specification of NSS, (6) \textit{NSS Instance to DL Translation}, where selected base translation rules along with an overview of the translation algorithm is provided, (7) \textit{Evaluation}, where we provide detailed analysis of experimental results of DLOL in comparison with existing OL tools on standard and proposed community datasets (8) \textit{Related Works} on OL approaches/tools and various OL submodules. 


\section{Preliminaries}
\label{preliminaries}
In this section, we introduce foundational concepts that will help to get better clarity of the problem statement and the approach sections to follow. We divide this section into three subparts. We first give a brief background on ontologies. Then, in subsequent section, we motivate the need of categorizing IS-A sentences. In the final section we give the formalism of $\mathcal{SHOQ}(\mathbf{D})$.
\subsection{Formal Knowledge Base}
In this section we give background formalisms, according to \cite{volker2007acquisition}, required to fully understand the problem statement.
\label{ontology def}
\begin{mydef}
	\textbf{Ontology}: An ontology is a structure $\mathcal{O}:=(C, \le_C, R, \sigma_R, \le_R, \mathcal{A}, \sigma_A, \mathcal{D})$ consisting of:
	\begin{itemize}
		\item four disjoint sets $C, R, \mathcal{A}, \mathcal{D}$ whose elements are called \textbf{concept} identifiers, \textbf{relation} identifiers, \textbf{attribute} identifiers and \textbf{datatypes}, respectively.
		\item a semi-upper lattice $\le_C$ on $C$, with top element $root_C$, called \textit{\textbf{concept hierarchy}} or \textit{\textbf{taxonomy}}.  
		\item a function $\sigma_R: R \ \mapsto C^+$ called \textbf{relation signature}; it captures the \textbf{domain} and \textbf{range} of a relation as a tuple. 
		\item a partial order $\le_R$ on R, called \textit{\textbf{relation hierarchy}}, where $r_1 \le r_2 \implies |\sigma_R(r_1)| = |\sigma_R(r_2)|$ and $\pi_i(\sigma_R(r_1)) \le_C \pi_i(\sigma_R(r_2))$ for each $1 \le i \le \sigma_R(r_1)$; where $\pi_i(t)$ is the i-th component of tuple \textit{t}.
		\item a function $\sigma_A: \mathcal{A} \ \mapsto C \times \mathcal{D}$ called \textbf{attribute signature}
		\item a set $\mathcal{D}$ of datatypes such as strings, integers, etc. 		
	\end{itemize}
\end{mydef}
\begin{mydef}
	\textbf{Knowledge Base}: A knowledge base (KB) for an ontology $ \mathcal{O} := (C,\le_C,R, \sigma_R, \le_R, \mathcal{A}, \sigma_A, \mathcal{D})$ is a structure: $KB := (I, l_C, l_R, l_A)$ consisting of:
	\begin{itemize}
		\item a set I whose elements are called \textbf{instance} identifiers (or \textbf{intances}, or \textbf{objects})
		\item a function $l_C: C \ \mapsto 2^{I}$ called \textbf{concept instantiation}.
		\item a function $l_R: R \ \mapsto 2^{I^+}$ with $l_R(r) \ \subseteq \ \Pi_{1 \le i \le | \sigma(r)|}\  l_C(\pi_i(\sigma(r)))$, for all $ r \in R$. The function $l_R$ is called \textbf{relation instantiation}, and
		\item a function $I_A:\ \mathcal{A} \to I \times \bigcup_{d \in \mathcal{D}}[\![d]\!]$ with $l_A(a) \subseteq \ l_C(\pi_1(\sigma(a))) \ \times \ [\![\pi_2(\sigma(a))]\!]$, where $[\![d]\!]$ are the values of datatype $ d\ \in \ \mathcal{D}$. The function $l_{A}$ is called \textbf{attribute instantiation}.
		
	\end{itemize}
\end{mydef}
\begin{mydef}
	\textbf{Extensions of Knowledge Base}: Let $KB := (I, l_C, l_R, l_A) $ be a knowledge base for an ontology $ \mathcal{O} := (C,\le_C,R, \sigma_R, \le_R, \mathcal{A}, \sigma_A, \mathcal{D})$. 
	\begin{enumerate}
		\item {The extension $[\![c]\!]_{KB} \ \subseteq \ I$ of a concept $c\ \in \ C$ is recursively defined by the following rules:			
			\begin{itemize}
				\item $[\![c]\!]_{KB} \leftarrow l_C(c)$
				\item $[\![c]\!]_{KB} \leftarrow [\![c]\!]_{KB} \cup [\![c']\!]_{KB}$, for $c' <_C c $.
				\item instantiations of axion schemata in S (if $\mathcal{O}$ is an ontology with $\mathcal{L}$-axioms),
				\item other general axioms contained in the logical theory.
			\end{itemize}}
		\item {The extension $[\![r]\!]_{KB} \ \subseteq \ I^+$ of a relation $r\ \in \ R$ is recursively defined in similar way as in \em{1}.}
		\item {The extension $[\![a]\!]_{KB} \ \subseteq \ I \times [\![\mathcal{D}]\!]$ of an attribute $a\ \in \ \mathcal{A}$ is recursively defined in similar way as in \em{1}.}
	\end{enumerate}

\end{mydef}
\begin{mydef}
	\textbf{Intension of Knowledge Base}: A structure $\Tau := (\mathcal{L}, i_C, i_R, i_\mathcal{A})$ is called an \textbf{intension} of an ontology $ \mathcal{O} := (C,\le_C,R, \sigma_R, \le_R, \mathcal{A}, \sigma_A, \mathcal{D})$ and consists of:
	\begin{itemize}
		\item a language $\mathcal{L}_I$ capturing intensions of conepts, relations, and attributes, respectively.
		\item three mappings $i_C, i_R, i_\mathcal{A}$ with $i_C : C \mapsto \mathcal{L}_I$, $i_R : R \mapsto \mathcal{L}_I$, and $i_\mathcal{A} : \mathcal{A} \mapsto \mathcal{L}_I$ 
	\end{itemize} 
	\emph{	Here, intension of a language refers to the formal semantics of interpreting any concept/relation/attribute definition or axiom (one of which is model-theory)}.
\end{mydef}
\begin{mydef}
	\textbf{Generalized Terminology}: A set of \textbf{axioms} $\mathcal{T}$ is a generalized terminology if the left-hand side of each axiom is an atomic concept and for every atomic concept there is at most one axiom where it occurs on the left-hand side.
\end{mydef}
\begin{mydef}
	\textbf{Regular Terminology}: A regular terminology is a set of \textbf{definitions} $\mathcal{\bar{T}}$. 
\end{mydef}
It is important to note that a generalized terminology can be converted to a regular terminology, and vice-versa (using a classification reasoner). 
\subsection{IS-A Sentences and Types}
\label{is-a type}
We now formalize our notion of IS-A sentences in English as follows:
\begin{mydef}
\textbf{IS-A Sentence}: If the predicate connective between subject/s and object/s of a sentence is either synonymous or similar to the phrase '\textit{is a kind of}' then the sentence is called an IS-A sentence.
\end{mydef}

For instance, ``\textit{John seems to be a financially sound person}" is an IS-A sentence. Here, the predicate connective ``\textit{seems to be}", between the subject ``\textit{John}" and the object ``\textit{a financially sound person}", is semantically similar to `\textit{is a kind of}'. We classify IS-A sentences in English into \textit{trivial} and \textit{non-trivial} sentences as follows:
\begin{mydef}
\textbf{Trivial IS-A Sentence}: An IS-A sentence is \textit{trivial} if its formal semantic construction, using the \textit{principle of compositionality}, is the \textit{correct} and \textit{complete} representation of the actual linguistic reading of the sentence. 
\end{mydef}

If a formal representation is correct then it should not add any additional semantics to the corresponding original natural language form. If a formal representation is complete then it should be able to capture every aspect of the linguistic semantics of the original natural language form. An example of a trivial IS-A sentence is ``\textit{John is a financially stable person}", which can be correctly and completely represented as a formal expression (say, First-Order Logic ), stating that ``\textit{John}" is an instance of the class\footnote{Class, in ontology parlance, is also called '\textit{concept}'.} ``\textit{financially stable person}".

\begin{mydef}
\textbf{Non-Trivial IS-A Sentence}: An IS-A sentence is \textit{non-trivial} if it has innate linguistic nuances that cannot be correctly or completely represented by conventional formal semantic construction, and thus, requires additional language-specific semantic construction-rules integration and modification.
\end{mydef}
 An example of a non-trivial IS-A  sentence is ``\textit{John seems to be a financially sound person}". Here, the relation ``\textit{seems to be}" is a particular variation of ``\textit{is a kind of}", implying that there is a possibility, although low, that \textit{John} may not be financially stable. We have identified 15 and 56 uniquely different cases of trivial and non-trivial IS-A sentences, respectively\footnote{See \crdurl}. Many IS-A sentences can structurally be quite complicated with more than one object and/or subject. This motivates a further classification of IS-A sentences into simple, complex, and compound IS-A sentences as follows:

\begin{mydef}
\textbf{Simple IS-A Sentence}: An IS-A sentence is \textit{simple} if it has only one subject, one object, and disallows clausal restrictions on both the subject and the object.
\end{mydef}
 An example simple sentence is: ``\textit{John is a financially sound person}".
 \begin{mydef}
\textbf{Complex IS-A Sentence}: An IS-A sentence is \textit{complex} if it has clausal restrictions, having \mbox{IS-A} structure, on the single subject or the single object\footnote{IS-A clauses are of the types ``\textit{that is ...}", ``\textit{which is ...}", ``\textit{being ...}", ``\textit{as ...}", etc.}.
 \end{mydef}
 An example of a complex IS-A type sentence is ``\textit{John, who is quite clever, is a financially sound person}", where  ``\textit{who is ...}" is of type IS-A.
 \begin{mydef}
\textbf{Compound IS-A Sentence}: An IS-A sentence is \textit{compound} if it has conjunctive or disjunctive list of subjects and/or objects.  \end{mydef}
A compound IS-A sentence can many times be simplified into either a set of simple IS-A type sentences, or that of complex sentences, or both. An example compound sentence is: ``\textit{John and Joe are financially sound.}"

\subsection{$\mathcal{SHOQ}(\bD)$: Formal Representation Language for Factual IS-A English Sentences}\label{DL}
We observed that most factual IS-A type sentences can be expressed using the DL fragment $\mathcal{SHOQ(\textbf{D})}$ \cite{63baader2007completing} where $\mathcal{S} \equiv \mathcal{ALC_{R^+}}$.\\

\begin{tabular}{p{1cm}p{14cm}}
$\mathcal{[R^+]}$ & Transitive Role $-$ used only for the relation \textbf{\textit{includes}}. An example of such usage is in the hypernymic IS-A sentence: ``\textit{Animals \textbf{include} marsupials and echidnae. Marsupials \textbf{include} kangaroos.}"\\
$\mathcal{[U]}$ & Union $-$ used only for some rare cases of entity phrase disjunction. An example of such sentence is: ``\textit{School can be \textbf{either a fun place or a boredom}.}"\footnotemark\\

$\mathcal{[C]}$ & Complement $-$ will not be used in positive IS-A type sentences.\\
$\mathcal{[O]}$ & Nominal $-$ will only be used during some cases of unrecognized named entity. An example sentence of usage is: ``\textit{\textbf{Priyansh} is a hard-working student.}", where \textit{Priyansh} is not recognized by any standard NER tool as a \textit{person}. Hence, \textit{Priyansh} has to be treated as a nominal.\\

$\mathcal{[H]}$ &  Role Hierarchy $-$ will only be used in association with entity conjunction together with the relation \textit{includes}. Details about of the case and justification of its usage is given in section \ref{DLNonTrivial}.\\
$\mathcal{[Q]}$ & Full Number Restriction $-$ will only be used during representing sentences having numeral subject entity quantifiers. An example usage case is: ``\textit{\textbf{At least three} candidates are qualified.}"\medskip
\end{tabular}

\footnotetext[8]{It is, however, to be noted that in most cases of English sentences disjunction, in the form of the lexicon \textit{or}, denotes conjunction. As an example, the sentence ``\textit{Students, such as John or Mary, are intelligent.}" has the same meaning as ``\textit{Students, such as John and Mary, are intelligent.}"}

The choice of DL over other semantic representation theories has several reasons: (i) DL is a decidable fragment of First-Order Logic \cite{baader2005description}, (ii) DL representation is compact, as compared to representations such as DRS \cite{75kamp1981theory} and LFT \cite{81moldovan2001logic} making it comparatively easy to parse, (iii) highly optimized semantic-tableau based reasoners are available, (iv) DL representation can be easily converted into the W3C recommended OWL format for web ontology\footnote{OWL DL is equivalent to $\mathcal{SHOIN(D)}$ while OWL 2 is equivalent $\mathcal{SROIQ(D)}$}. Also, since DL is a variable-free representation, certain linguistic issues, such as resolving ``\textit{donkey sentences}"\footnote{A \textit{donkey sentence} contains a special kind of linguistic \textit{anaphora} that is bound in semantics but not in syntax. This leads to inadequate representation of such sentences in FOPL and requires more elaborate techniques such as DRS. An example sentence is: ``\textit{Every farmer, who owns a donkey, beats it.}", where a montague-style semantic construction can lead to two readings: (i) each donkey owning farmer beats his donkey/s, and (ii) every donkey is beaten by each donkey owning farmer.}, becomes much easier. Expressions in DL can represent two types of sentences: (i) general facts such as ``\textit{Some person are students}" (technically called \textit{T-Box definitions/inclusion axioms}), and (ii) specific ground facts such as ``\textit{John is a student}" (technically called \textit{A-Box assertions}). Every A-Box assertion must have a corresponding T-Box induction for maintaining ontology consistency. We briefly recall the description logic (DL) $\mathcal{SHOQ}(\bD)$, 
which turns out to suit very well our approach to OL.\\

\newcommand{\KB}{K\!B}
\textbf{Syntax:} We assume a set $\bD$ of \textit{concrete datatypes}. Each datatype $d\ins\bD$ has a {\em domain}~${\rm dom}(d)$. We use ${\rm dom}(\bD)$ to denote the set of all {\em (data) values}, which is the union of the domains ${\rm dom}(d)$ of all datatypes~$d\ins\bD$. Let $\mathbf{C}$, $\mathbf{R}_A$, $\mathbf{R}_D$, and~$\mathbf{I}$ be nonempty finite disjoint sets of {\em atomic concepts}, {\em abstract roles}, {\em concrete roles}, and {\em individuals}, respectively. \textit{Concepts} are inductively defined as follows:
\begin{itemize}\setlength{\itemsep}{0.25ex}
	\item Every atomic concept from~$\mathbf{C}$ is a concept. 
	\item If $o$ is an individual from $\mathbf{I}$, then $\{o\}$ is a concept. 
	\item If $C$ and $D$ are concepts, then so are $(C\sqcap D)$, $(C\sqcup D)$, and $\neg C$ (called {\em conjunction}, {\em disjunction}, and {\em negation}, respectively). 
	\item If $C$ is a concept, $R\ins \mathbf{R}_A$, and $n$ is a nonnegative integer, then $\exists R.C$, $\forall R.C$, ${\ge} n R.C$, and ${\le} n R.C$ are concepts ({\em existential}, {\em value}, {\em atleast}, and {\em atmost restriction}, respectively).
	\item If $T\ins\mathbf{R}_D$, and $d$ is a concrete datatype
	from $\mathbf{D}$, then~$\exists T.d$ and $\forall T.d$ are concepts ({\em datatype existential} and {\em value restriction}, respectively). 
	\item We write $\top$ (resp., $\bot$) to abbreviate $C\sqcup \neg C$ (resp., $C\sqcap \neg C$), and we eliminate parentheses as usual. 
\end{itemize}

A {\em concept inclusion axiom} is an expression $C\sqsubseteq D$, 
where $C$ and $D$ are concepts. A {\em role inclusion axiom} is an expression of the kind $R\sqsubseteq S$, where either 
$R,S\ins \mathbf{R}_A$ or~$R,S\ins \mathbf{R}_D$. A~{\em transitivity 
	axiom} has the form ${\rm Trans}(R)$, where $R\ins \mathbf{R}_A$.
A~{\em concept membership axiom} is of the form $C(o)$, 
where $C$ is a concept, and $o$ is an individual. 
An {\em abstract role membership axiom} is of the form $R(o_1,o_2)$, 
where $R$ is an abstract role, and $o_1$ and $o_2$ are individuals. 
A {\em concrete role membership axiom} is of the form $T(o,v)$, 
where $T$ is a concrete role, $o$ is an individual, and $v$ is a value. 
A {\em terminological axiom} is either a concept inclusion axiom, 
a role inclusion axiom, or a transitivity axiom. An {\em assertional axiom} is either a concept membership axiom, 
an abstract role membership axiom, or a concrete role membership axiom.
A {\em TBox} $\cT$ is a finite set of terminological axioms.
An {\em ABox} $\mathcal{A}$ is a finite set of assertional axioms.
A {\em knowledge base} (or {\em ontology}) $\KB=(\mathcal{T},\mathcal{A})$ consists of a TBox $\mathcal{T}$ and an ABox $\mathcal{A}$. \\

\textbf{Semantics:} An {\em interpretation} ${\cal I}\eqs (\Delta,I)$ relative to $\mathbf{D}$ consists of a nonempty {\em (abstract) domain} $\Delta$ and a mapping~$I$ that assigns
to each atomic concept from $\mathbf{C}$ a subset of $\Delta$, 
to each $o\ins \mathbf{I}$ an element of $\Delta$, 
to each abstract role from~$\mathbf{R}_A$ 
a subset of $\Delta\times \Delta$, 
and to each concrete role from~$\mathbf{R}_D$
a subset of  $\Delta\times {\rm dom}(\mathbf{D})$.
The interpretation $I$ is extended by induction
to all concepts as follows (where $\# S$ 
denotes the cardinality of a set $S$):
\begin{itemize}\setlength{\itemsep}{0.25ex}
	\item $I(C\sqcap D)=I(C)\cap I(D)$, $I(C\sqcup D)=I(C)\cup I(D)$, 
	and $I(\neg C)=\Delta\,{\backslash}\, I(C)$,
	\item[$\bullet$] $I(\exists R.C)=\{x\ins \Delta\mids \exists y\colon (x,y)\ins
	I(R)\wedge y\ins I(C)\}$,
	\item[$\bullet$] $I(\forall R.C)=\{x\ins \Delta\mids \forall y\colon (x,y)\ins
	I(R)\rightarrow y\ins I(C)\}$,
	\item[$\bullet$] $I({\ge} n R.C)=\{x\ins \Delta| \# (\{y\mid
	(x,y)\ins I(R)\}\caps I(C))\ges n\}$,
	\item[$\bullet$] $I({\le}n R.C)\eqs\{x\ins \Delta| \# (\{y\mid
	(x,y)\ins I(R)\}\caps I(C))\les n\}$,
	\item[$\bullet$] $I(\exists T.d)=\{x\ins \Delta\mids \exists y\colon (x,y)\ins
	I(T)\wedge y\ins {\rm dom}(d)\}$, 
	\item[$\bullet$] $I(\forall T.d)=\{x\ins \Delta\mids \forall y\colon (x,y)\ins
	I(T)\rightarrow y\ins {\rm dom}(d)\}$.
\end{itemize}
The {\em satisfaction} of an axiom $F$ 
in ${\cal I}$, denoted $\cI\models F$, is defined by: 
(i) $\cI\models C\sqsubseteqs D$ iff $I(C)\subseteqs I(D)$,
(ii) $\cI\models R\sqsubseteqs S$ iff $I(R)\subseteqs I(S)$,
(iii) $\cI\models {\rm Trans}(R)$ iff $I(R)$ is transitive, 
(iv) $\cI\models C(o)$ iff $I(o)\ins I(C)$, 
(v) $\cI\models R(o_1,o_2)$ iff $(I(o_1),$ $I(o_2))\ins I(R)$,
and (vi)  $\cI\models T(o,v)$ iff $(I(o),v)\in I(T)$.
The interpretation ${\cal I}$ {\em satisfies} the axiom 
$F$, or ${\cal I}$ is a {\em model} of $F$, 
iff $\cI\modelss F$. It {\em satisfies} a knowledge base $\KB\eqs(\mathcal{T},\mathcal{A})$, 
or ${\cal I}$ is a {\em model} of $\KB$, denoted ${\cal I}\modelss \KB$, iff 
${\cal I}\modelss F$ for all~$F\ins \mathcal{T}\cups\mathcal{A}$. We say $\KB$ is 
{\em satisfiable} iff $\KB$ has a model. An axiom 
$F$ is a {\em logical consequence} of $\KB$, denoted $\KB\modelss F$, 
iff every model of $\KB$ is also a model of~$F$.

\section{Problem Statement} 

\subsection{Significance}
One of the key milestones of modern AI is to build intelligent applications that mimic human-like understanding of textual (i.e. natural language) data. Many contemporary software systems benefit from ontology learning on texts. To cite a few, Google \footnote{\url{http://ok-google.io/}}, Microsoft \footnote{\url{http://www.windowscentral.com/cortana}}, and Apple \footnote{\url{http://www.apple.com/in/ios/siri/}} have their in-house question answering systems and digital assistants. Their ability to answer varied and extensive questions, in a large way, depends on the availability of comprehensive machine-readable knowledge. In most scenarios this knowledge is represented as generic conceptual networks \cite{liu2004conceptnet, dong2014knowledge}, lexical taxonomies \cite{79miller2005wordnet}, and domain-specific ontologies \cite{bodenreider2004unified, rogers1996galen}. As mentioned in the introduction, ontology engineering is difficult and labor intensive. Furthermore, ontologies shared among different applications, organizations, and individuals, need to agree on common design choices to maintain meaningful and consistent generalization. Hence an ideal solution would be a scalable high performance and accurate OL framework.

\subsection{Formal OL on Text: Motivation \& Definition}
We first define the problem of formal OL in the context of IS-A sentences in English as follows:
\begin{mydef}
	\textbf{Formal OL on Text}: Given a corpus (of IS-A sentences, as per the scope of this paper) in English, formal OL is the process of automatically generating a consistent ontology that represents the corpus, as a knowledge base, comprising of its intension (as generalized and regular terminology) and its extension, in some chosen target formal language.
\end{mydef}
The motivation to model the problem as formal OL, as opposed to lightweight OL, can be summed up by the following rationale:
\begin{enumerate}
	\item Formal ontologies (i.e. regular and generalized terminologies) have an intentional knowledge base (also called \textit{T-Box}), which can be used for abstract reasoning on entities and relations, classification based on relations between entities, as well as detecting contradiction and inconsistencies between them.
	\item At the same time, the existence of extensional knowledge base (also called \textit{A-Box}) enables retrieval of implicit abstractions of facts (also called \textit{assertions}), and their classification.
\end{enumerate}

Formal OL on IS-A sentences is quite challenging from a linguistic analysis point of view. The non-triviality of \mbox{IS-A} sentences is discussed in depth in \cite{10brachman1983and,11krifka2013definitional}. This is evident from the number of different types of factual IS-A sentences\footnote{Available in \crdurl}. Moreover, we believe that a system which aims at performing micro-reading on text to perform deep semantic analysis of sentences \cite{nell_2015} will benefit as IS-A sentences form basic building block for more semantically rich sentences.

\section{Approach Overview}
\label{approach}
\subsection{Motivation}
\label{motivation}
Natural Language (NL) sentences can be linguistically modeled as \textit{triples}, comprising of \textit{subject} (termed \textit{S}), \textit{relation} (termed \textit{R} or \textit{V}), and \textit{object} (termed \textit{O})\footnote{The SRO modeling is in accordance with the field of \textit{linguistic typology} (see \url{https://en.wikipedia.org/wiki/Linguistic\_typology})}. The subject is a noun phrase while the relation together with the object forms a verb phrase (which itself can be broken recursively into a verb phrase representing the relation and noun phrase representing the object). In Description Logics (DL) representation of NL sentences, the subject of a sentence can be modeled as \textit{derived concept}, which is ``\textit{defined}" (in the definitorial sense) or ``\textit{related}" (in the axiomatic sense) in terms of its relations (i.e. \textit{roles}) and their associations with objects (i.e. \textit{filler concepts}). Therefore, identification of ``\textit{subject-object dependency}" in a given sentence is pivotal in extracting these triples. As an example, in the sentence ``\textit{John, who is a student, is also a musician}", both the objects \textit{student} and \textit{musician} are associated with the subject \textit{John}. Here, two triples can be extracted: ``\textit{John is a student}"; and ``\textit{John is a musician}". It can be observed that the core grammatical structure of an IS-A sentence is of the form: $\textit{Subject} \quad \textit{IS-A} \quad \textit{Object}$. It is in this direction that we propose a chunker-styled ``\textit{pseudo-grammar}" template called \textit{Normalized Sentence Structure} (\textit{NSS}). NSS helps to  restructure sentences, having different syntactic variations, into a common normalized ``\textit{template form}". In this way, NSS serves as an intermediary template language to translate a sentence into its corresponding DL expression using a generic translation algorithm. However, restructuring of IS-A sentences may be quite complex and hence, may require several text pre-processing normalization steps. Also, there are several intrinsic linguistic nuances involved in DL translation that also require certain specific pre-processing. We briefly describe each such processing modules in the following sections.

\begin{figure*}[!t]
	\centering
	\includegraphics[scale=0.40]{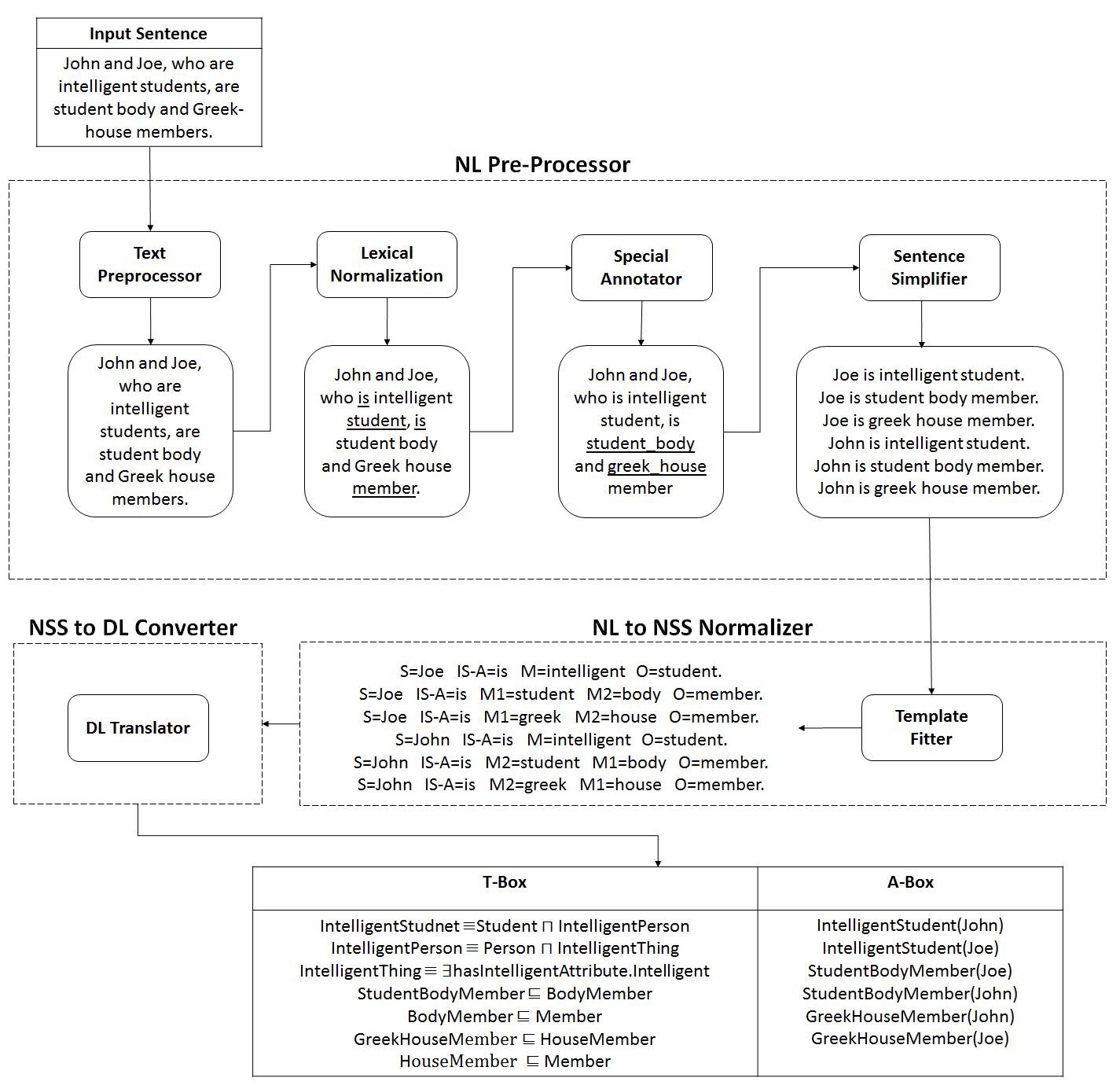}
	\caption{DLOL Architecture Pipeline}
\label{pipeline}
\end{figure*}

\subsection{DLOL Architectural Pipeline}
\label{architecture}
The DLOL system pipeline has three principal components: (i) \textit{NL Pre-Processor}, (ii) \textit{NL to NSS Normalizer}, and (iii) \textit{NSS to DL Translator}. We discuss each of them as follows:\\

\textbf{I. NL Pre-Processor}: This is the first component and is responsible for annotating and simplifying the IS-A English sentences. We employ various NLP techniques to make these sentences suitable for easy normalization into NSS instances. 

\textbf{II. NL to NSS Normalizer}: This component takes its input from the NL Pre-processor, and  the processed sentences into an intermediary canonical representational structure (NSS Instances). NSS fitting algorithm, which is a part of this module, fits the sentences into instances of some pre-defined templates.

\textbf{III. NSS To DL Converter}: This is the final module where an NSS instance (i.e. a sentence fitted into an NSS template) gets translated into a DL expression. The general methodology is to parse the various chunks of the NSS instance, invoke corresponding DL translation rules (depending upon the way the NSS template is fitted), and then compose the translations into the final output expression. More details are given in section \ref{NSS2DL}.
\subsubsection{DLOL NLP Modules}
\label{DLOL Modules}
The NL Pre-Processor has four core modules: (i) \textit{Text Pre-Processor}, (ii) \textit{Lexical Normalizer} (iii) \textit{Special Annotator}, and (iv) \textit{Sentence Simplifier}. We perform these tasks in order to make it easier to parse these sentences in the subsequent component, while keeping their semantic content intact.\\

\textbf{I. Text Pre-Processor}: This is the first module where preliminary pre-processing of text takes place. This includes various sub-modules like spell-corrector (so as to eliminate incorrect addition of “mispelled new concepts” in the ontology), tokenizer, sentence boundary detector, text singularizer, POS-tagger, co-reference resolver, and named-entity recognizer. First, we run the spell-corrector on the entire corpus, followed by tokenization, POS-tagging and sentence boundary detection. At the end of this process, the entire corpus is transformed into a list of sentences, and each sentence into a list of POS-tagged word-tokens. This is followed by annotating all the entities present in text with the help of a named entity recognizer system. Further, we try to find all the expressions which refer to same entity across the/different sentences by the co-reference resolver module. Finally, in the text singularizer module, we normalize the singular-plural variations of text. For example, ``\textit{The men are hardworking}" is converted into ``\textit{Man$_{\emph{PLURAL}}$ is hardworking}". Even though we normalize these variations, we keep the plural tag ``PLURAL" along to avoid representation error due to tense disagreement at later stage. Except for spell-correction, which we developed, all other tools are part of the Stanford coreNLP toolkit\footnote{\url{http://nlp.stanford.edu/software/corenlp.shtml}}.\\

\textbf{II Lexical Normalization}: This is the second module, where all the sentences extracted by the Text Pre-Processor are normalized lexically. At this stage, all lexical variations of \textit{is a kind of} and other linguistic elements, such as quantifiers of subject/object, and clauses of complex sentences, are identified. Each lexical variation present in the sentence is translated into their corresponding pre-determined normal forms (i.e. \textit{is} for IS-A lexemes, \textit{some} for quantifiers, etc.). All numeral lexemes are also converted into their corresponding number format to ensure standardized and efficient DL translation. As an example, “one hundred and thirty seven” will be converted into “137”. We used a WordNet \cite{79miller2005wordnet} based semi-automatic bootstrapping technique for hash-storing all lexemes that need to be normalized. Apart from specific lexeme normalization, all noun phrases are also normalized into their corresponding pre-determined synonymous forms.\\ 

\interfootnotelinepenalty=10000
\textbf{III. Special Annotator:} This module adds additional tags to lexemes for better DL representation of sentences. There are three different tasks that the module performs. 
First, it identifies dimensional lexemes, called \textit{units}, and mapping them to their corresponding dimensions. Units are tokens that correspond to quantitative measure for a dimension. For example, in the sentence ``\textit{John is five feet tall}", the unit ``\textit{feet}" is a measure for the dimension \textit{Height}. In total 19 core units have been identified, a few of them being \textit{foot}, \textit{kilogram}, \textit{bits}, \textit{meter}, \textit{year}, \textit{money}, \textit{degree}, \textit{meter per second}, \textit{second}~\footnote{Sub-types of a particular unit such as \textit{centimeter} have been separately bootstrapped}. We have identified 17 quantifiable dimensions\footnote{The seventeen quantifiable dimensions are \textit{Height}, \textit{Length}, \textit{Breadth}, \textit{Depth}, \textit{Weight}, \textit{Digital Size}, \textit{Area}, \textit{Volume}, \textit{Time}, \textit{Date}, \textit{Age}, \textit{Speed}, \textit{Acceleration}, \textit{Distance}, \textit{Currency}, \textit{Physics}, and \textit{Cardinality}.  The \textit{Physics} dimension may have sub-classification such as \textit{Energy}, \textit{Voltage}, \textit{Pressure}, etc. (Semi)-automatic bootstrapping of such specific dimensions can be a very interesting study on its own right since it can support several domain-specific reasoning during knowledge discovery. All the seventeen dimensions have a parent concept named \textit{Dimension}.}. All lexemes that are identified as units are annotated along with POS Tag as POS Tag\_UNIT\_Dimension. For example, the lexeme \textit{feet} is tagged as \textit{feet\_NNS\_UNIT\_HEIGHT}. \\

\textbf{IV. Sentence Simplifier}: The primary task of this module is to normalize complex and compound sentences into a set of further simplified sentences (see section \ref{preliminaries}). For example, the compound sentence: ``\textit{John and Joe, who are intelligent students, are student body and Greek-house members}" can be normalized into the following set of pure complex sentences: (i) ``\textit{John, who is intelligent student, is a student body member}", (ii) ``\textit{John, who is intelligent student, is Greek-house member}", (iii) ``\textit{Joe, who is intelligent student, is student body member}", and (iv) ``\textit{Joe, who is intelligent student, is Greek-house members}". All these four sentences can be further simplified into simple sentences. For example, sentence (i) can be simplified into: (i.a) ``\textit{John is intelligent student}”, and (i.b) ``\textit{John is student body member}". Hence, the original sentence gets simplified into six distinct simple sentences. However, if subject or object of a sentence are in disjunction then the resulting simplified sentences can be a set of simpler complex/compound sentences. For an example, a variation of the previous sentence: ``\textit{Either John or Joe, who is a good student, is a student body member.}" will be simplified into two compound sentences: (i) ``\textit{Either John or Joe is a good student}”, and (ii) ``\textit{Either John or Joe is a student body member}". 

This module also re-structures (i.e. paraphrases) sentences into NSS template (see section \ref{motivation}) through a process called \textit{syntactic normalization}. For instance, the sentence: ``\textit{John's name is popular}" gets normalized to ``\textit{John is a popular name}". We have identified 47 lexico-syntactic pattern-rules where normalization is needed. These rules were independently framed on cases coming through (a) study of various books on linguistic semantics, (b) discussion with a professional linguist, (c) brainstorming sessions within the group.\footnote{The full list is available at \snrurl}. Examples of selected syntactic patterns are shown in Table \ref{tab:stmt_simple_example}. 

\begin{table}[!t]
	\caption{Selected 7 out of 47 syntactic patterns  with example to demonstrate the task of sentence simplification. Original syntax indicates a given sentence which is simplified to handle syntactic and lexical variations. }
	\label{tab:stmt_simple_example}
	\smallskip 
\begin{tabular}{p{4.5cm}p{5cm}p{6cm}}
	\hline
	Original syntax & Simplifies to & Example \\
	\hline \\	
	$NN_1$ \textit{is category of} $NN_2/NNP$ & $NN_2/NNP$ \textit{is a} $NN_1$ & ``\textit{Fruit is the category of apple}" \newline simplifies to:\newline ``\textit{Apple is a fruit}'' \newline \\
	$NN_1$ belongs to the group of $NN_2$ & $NN_1$ is a $NN_2$  
			&``Thyroid medicines comes under the \newline general group of hormone medicines." \newline simplifies to :\newline ``Thyroid medicine is a hormone medicine." \newline \\
	Some examples of $NN_1$ are $NN_2$, $NN_3$ and $NN_n$ & $NN_i$ is a $NN_1$ where $2 \leq i \leq n$	& ``A few examples of peripherals are keyboards,  scanner, printer and tape drive."
	\newline simplifies to:\newline	Key board is a peripheral.
	 \newline	Scanner is a peripheral. 
	\newline	Printer is a peripheral. 
	\newline	Tape drive is a peripheral. \newline \\

    There are $CD$ types of $NN_1$ : $NN_2, \ NN_3, \dots $ and $NN_{CD}$ & $NN_i$ is a $NN_1$ where  $2 \leq i \leq CD$ & ``There are two kinds of narcotic analgesics \newline - the opiates and the opioids." 
    \newline simplifies to : \newline Opiates is a narcotic analgesics
    \newline Opioids is a narcotic analgesics \newline \\
    
   It looks like $NN/NNP$ is $DT \ NN/JJ$ & $NN/NNP$ may be $DT \ NN/JJ$. & ``It seems that John is a good student." \newline simplifies to:\newline ``John may be a good student." \newline \\
	$NN_1$ and $NN_2$ are similar & $NN_1$ is similar to $NN_2$. & ``Tangerine and orange are alike." \newline simplifies to:\newline ``Tangerine is similar to orange." \newline \\
	$NNP/NN_1$ is $JJR \ NN_2$ & $NNP/NN_1$ is $JJR$ than some $NN_2$. & ``John is a taller player." \newline simplifies to:\newline ``John is taller than some player." \\
	\hline 
\end{tabular}
\end{table}
It is interesting to note that, although semantic construction of IS-A sentences is challenging, yet paraphrasing of such sentences can be captured by a set of fixed pattern-rules only (without having to resort to sophisticated lexico syntactic machine-learning techniques). 

\section{NL Sentence to NSS Instance}
\subsection{NSS Syntax}
NSS identifies subject/object phrases as chunks having the core structure: $\langle$DT, NP$\rangle$\footnote{DT: Determiner (ex: the, a, an, any, some, etc.); NP: Noun Phrase; (notations according Penn Tree Bank \cite{78marcus1993building}). All POS notations in this paper will follow the same convention} using a POS-tagger. Some examples of chunks are: ``\textit{a cat}'', ``\textit{the clever man}'', ``\textit{every student}'', etc. The predicates linking the subject-object phrases are also identified as chunks. In case of IS-A type sentences, there will be some lexical variation of: (i) \textit{is a type of} (subject hyponymy), (ii) \textit{is the category of} (subject hypernymy), (iii) \textit{is similar to} (subject-object kinship), and (iv) \textit{is same as} (subject-object equivalence). In this way, NSS avoids unnecessary re-construction of the subject/object phrases over a parse tree\footnote{It is to be noted that for more complex non IS-A structures NSS uses dependency grammar, the discussion of which is out of the scope of this paper.}. As mentioned in section \ref{motivation}, the core NSS template for all IS-A type sentences is $\textit{S} \quad \textit{IS-A} \quad \textit{O}$ where \textit{S} denotes \textit{Subject}, \textit{O} denotes \textit{Object}, and \textit{IS-A} denotes a set of lexicons that are IS-A type variations. 
A \textbf{simple IS-A} sentence (defined in previous section) has the NSS template : \\

\begin{minipage}{\textwidth}
\begin{center}
$\dashuline{Q_{1}} \hspace{1em} \dashuline{M_{1}}^{*} \hspace{1em} S \hspace{1em} IS-A \hspace{1em} \dashuline{Q_{2}} \hspace{1em} \dashuline{M_{2}}^{*}  \hspace{1em} O$\\
\end{center}
\end{minipage}

\noindent where:\\

\begin{tabular}{cl}
$\dashuline{\hspace{0.2cm}Q\hspace{0.2cm}}$ & Dashed line indicates optional component with at most 1 occurrence in the template e.g. a quantification.\\
$\dashuline{\hspace{0.2cm}M^{*}\hspace{0.1cm}}$ & Dashed line with asterisk (*) indicates 0 or more consecutive occurrence of a component in the template e.g. adjectives.\\
$Q_{1}$ & Subject quantifier that includes lexical variations of the set: \{\textit{a, an, the, some, all}\}. \\
$Q_{2}$ & Object quantifier that includes lexical variations of the set: \{\textit{the, some, all}\}.\\
$M$ & Subject/object/verb modifier; value is restricted to the set: \{NN, JJ, CD, RB, VBG\}\footnotemark\\
$S$ & Subject; value is restricted to the set: \{NN, NNP\footnotemark, JJ, RB, VBG\}\\
$O$ & Object; value is restricted to the set: \{NN, NNP, JJ, RB, VBG\}\\
$\textit{IS-A}$ & Denotes all possible lexical variations. \medskip
\end{tabular}

\footnotetext[22]{NN: Noun; JJ: Adjective; RB: Adverb; VBG: Gerund; CD: Numeral/Cardinal (notations according to Penn Treebank)}
\footnotetext[23]{NNP: Proper Noun according to Penn Treebank.}
\noindent An example NSS instance for the simple sentence ``\textit{The dog is very clever}" is\\

\begin{minipage}{\textwidth}
\centering
\begin{tabular}{ccccccc}
Q$_{1}$ = \textit{The}& & S = \textit{dog} & IS-A = \textit{is}& & M$_2$ = \textit{very} & O = \textit{clever} \\
 & M$_1$ = \textit{null} & & & Q$_{2}$ = \textit{null} & & 
\end{tabular}
\end{minipage}  


\pagebreak

\noindent In a similar manner, the NSS template for a \textbf{complex IS-A} sentence is \\
\begin{center}
\begin{minipage}{\textwidth}
\begin{center}
$
\dashuline{Q_{1}} \hspace{1em} \dashuline{M_{1}}^{*} \hspace{1em} S
\hspace{1em} Cl_{1} \hspace{1em} IS-A \hspace{1em}
\dashuline{Q_{2}} \hspace{1em} \dashuline{M_{2}}^{*} \hspace{1em} O_{1}
\hspace{1em} Cl_{2} \hspace{1em} IS-A \hspace{1em}
\dashuline{Q_{3}} \hspace{1em} \dashuline{M_{3}}^{*} \hspace{1em} O_{2}$
\end{center}

\noindent where:\\

\begin{tabular}{cl}
$\dashuline{\hspace{0.2cm}Q\hspace{0.2cm}}$ & Dashed line indicates optional component with at most 1 occurrence in the template e.g. a quantification.\\
$\dashuline{\hspace{0.2cm}M^{*}\hspace{0.1cm}}$ & Dashed line with asterisk (*) indicates 0 or more consecutive occurrence in the template e.g. adjectives.\\
$Q_{1}$ & Subject quantifier that includes lexical variations of the set: \{\textit{a, an, the, some, all}\}. \\
$Q_{2}$ & Object quantifier that includes lexical variations of the set: \{\textit{the, some, all}\}.\\
$Q_{3}$ & Object quantifier that includes lexical variations of the set: \{\textit{the, some, all}\}.\\
$M$ & Subject/object/verb modifier; value is restricted to the set: \{NN, JJ, CD, RB, VBG\}\footnotemark\\
$S$ & Subject; value is restricted to the set: \{NN, NNP\footnotemark, JJ, RB, VBG\}\\
$O$ & Object; value is restricted to the set: \{NN, NNP, JJ, RB, VBG\}\\
$\textit{IS-A}$ & Denotes all possible lexical variations. \\
$Cl_{i}$ & signifies IS-A clausal token and all its variations \textit{`which', `who', `whose', `whom', `that'}\medskip
\end{tabular}

\begin{tabular}{cl}

\end{tabular}

\end{minipage}
\end{center}

\footnotetext[24]{NN: Noun; JJ: Adjective; RB: Adverb; VBG: Gerund; CD: Numeral/Cardinal (notations according to Penn Treebank)}
\footnotetext[25]{NNP: Proper Noun according to Penn Treebank.}

Note that in this template both $Cl_{1}$ (called the \textit{subject clause}) and $Cl_{2}$ (called the \textit{object clause}) cannot co-exist since that will render the sentence unsaturated. There are certain clausal tokens like \textit{`being', `seemingly', `looking', `having been', `as'} etc. that have null $Cl$ component. For example, the complex sentence: ``\textit{John, being a hard-working student, is successful}" is characterized as:

\begin{center}
\begin{minipage}{\textwidth}

S=\textit{John} \hspace{0.5cm} IS-A=\textit{being} \hspace{0.0cm} $Q_2$=a \hspace{0.0cm} $ M_{2}$=\textit{hard-working} \hspace{0.0cm} O$_{1}$=\textit{student} \hspace{0.8cm} IS-A=\textit{is}  \hspace{2.5cm} O$_{2}$=\textit{successful}\linebreak
\indent \hspace{0.7cm} Cl$_{1}$=\textit{null} \hspace{7.0cm} Cl$_{2}$ = \textit{null} \hspace{0.8cm} Q$_{3}$ = \textit{null} \hspace{0.0cm} M$_{3}$ = \textit{null}\\

\end{minipage}
\end{center}


\noindent The NSS instance for a base compound sentence is:\\

\begin{minipage}{\textwidth}
$\dashuline{Q_{1}} \hspace{1em} \dashuline{M_{1}}^{*} \hspace{1em} \wedge S_{1..k}/\vee S_{1..k}  \hspace{1em} \dashuline{Cl_{1}}  \hspace{1em} IS-A  \hspace{1em}  \dashuline{Q_{2}} \hspace{1em} \dashuline{M_{2}}^{*}  \hspace{1em}  \wedge O^{1}_{1..l}/\vee O^{1}_{1..l} \hspace{1em}  \dashuline{Cl_{2}} \hspace{1em} IS-A  \hspace{1em} \dashuline{Q_{3}} \hspace{1em} \dashuline{M_{3}}^{*}  \hspace{1em}  \wedge O^{2}_{1..m}/\vee O^{2}_{1..m} $\\
\newline
where:\\
\newline
\begin{tabular}{cp{14cm}}

$\wedge S_{1..k}/\vee S_{1..k}$ & Conjunctive/Disjunctive list of subject.\\
$\wedge O^1_{1..k}/\vee O^1_{1..k}$ & Conjunctive/Disjunctive list of first object.\\
$\wedge O^2_{1..k}/\vee O^2_{1..k}$ & Conjunctive/Disjunctive list of second object.\\
remaining symbols & have same semantics as above. \medskip
\medskip
\end{tabular}
\end{minipage}


\noindent More complicated sentences can be recursively fitted into templates such as:
\begin{center}
\begin{minipage}{\textwidth}
$\dashuline{Q_{1}} \hspace{0.3em} \dashuline{M_{1}}^{*} \hspace{0.3em} \wedge S_{1..k}/\vee S_{1..k} \hspace{0.3em}  \dashuline{\hspace{0.5em}\vee / \wedge  \hs \wedge S_{k..n} /\vee S_{k..n} \hspace{0.5em}} \hspace{0.3em} \dashuline{Cl_{1}}  \hspace{0.4em} IS-A  \hspace{0.4em}  \dashuline{Q_{2}} \hspace{0.4em} \dashuline{M_{2}}^{*} \hspace{0.4em}  \wedge O^{1}_{1..l}/\vee O^{1}_{1..l} \hspace{0.4em}  \dashuline{Cl_{2}} \hspace{0.4em} IS-A  \hspace{0.4em} \dashuline{Q_{3}} \hspace{0.4em} \dashuline{M_{3}}^{*}  \hspace{0.4em}  \wedge O^{2}_{1..m}/\vee O^{2}_{1..m} $\\
\end{minipage}
\end{center}
NSS template differs from Hearst-like patterns \cite{47hearst1992automatic} in ways such as: (i) NSS instance can be directly translated into a corresponding DL expression, and (ii) unlike Hearst-like patterns, NSS helps in generating a common DL expression for sentences that may have equivalent semantic content but different syntactic structures. 

\begin{algorithm*}[t]
	\KwData{$S_{\textit{PRE-PROCESSED}}, \textit{QUANTIFIER\_LIST}, \textit{IS-A\_LIST}$}
	\KwResult{NSS Instance}
	\Begin{
		$ARRAYLIST  \ \textit{SimplifiedSentenceSet} \longleftarrow SIMPLIFY(S)$\;
		\For{$\textit{sentence} \in \textit{SimplifiedSentenceSet}$}
		{
			$\textit{is-a\_lexeme} \longleftarrow \textit{EXTRACT\_IS-A}(\textit{sentence}, \textit{IS-A\_LIST})$\;
			$INSERT\_IN\_NSS([\textit{IS-A}], \textit{is-a\_lexeme})$\;
			\smallskip
			$\textit{subject\_phrase} \longleftarrow			
			\textit{EXTRACT\_BEFORE}(\textit{sentence}, \textit{is-a\_lexeme})$\;
			\For{$\textit{token} \in \textit{subject\_phrase}$}
			{
				\If{$\textit{token} \in \textit{QUANTIFIER\_LIST}$}
				{
					$INSERT\_IN\_NSS([\textit{Q}], \textit{token})$\;
				}
				\ElseIf{$EXTRACT\_POS-TAG(\textit{token})\in \{NN, NNP, JJ, VBG, RB, CD\}$}
				{
					\While{$NEXT\_TOKEN(token) \neq \textit{is-a\_lexeme}$}
					{
						$\textit{token} \longleftarrow EXTRACT\_NEXT\_TOKEN(token)$\;
						\If{$EXTRACT\_POS-TAG(\textit{token})\in \{NN, NNP, JJ, VBG\}$}
						{
							$INSERT\_IN\_NSS([M_S],\textit{token})$\;
						}
					}
					$INSERT\_IN\_NSS([\textit{S}], \textit{token})$\;	
				}						
			}
			$\textit{object\_phrase} \longleftarrow \textit{EXTRACT\_AFTER}(\textit{sentence}, \textit{is-a\_lexeme})$\;
			(Same steps repeated for $object\_phrase$ as in $subject\_phrase$)
		}
	}
	\caption{NSS Template-Fitting Algorithm \label{TemplateFitting}}
\end{algorithm*}

\subsection{NSS Template-Fitting Algorithm}
\label{Template Fitting Algorithm}
We now describe the template-fitting algorithm here. A sentence first undergoes a few basic NLP pre-processing (as discussed in \ref{architecture}). It is then fed into the template fitter, which already has a list of identified quantifier variations (i.e. \textit{some} variations) and IS-A variations (i.e. \textit{is} variations). Then subject phrase and object phrase are extracted by identifying the IS-A lexeme in the sentence. After that, tokens in these phrases are inserted into the NSS cells by identifying their corresponding POS tags. The outline of the algorithm is given in Algorithm \ref{TemplateFitting}.

\section{NSS Instance to DL Translation}
\label{NSS2DL}

\subsection{Selected DL Translation Rules}
Depending on the type of IS-A sentences, DL translation rules are categorized into two category (i) \textit{trivial DL translation rules} and (ii) \textit{non-trivial DL translation rules}.
\subsubsection{Trivial DL Translation Rules}
\label{DLTrivial}
In this section we consider particular cases of trivial IS-A sentences, that can be generalized as: (i) sentence with inclusion, (ii) sentence with modifiers, and (iii) sentence with subject quantifiers.\\

\noindent\textbf{I. Trivial Inclusion Rule:}
\label{trivialInclusion}
One of the most basic cases in this category is the following NSS:
\begin{enumerate}
\item { S=NNP \footnote{S=NNP includes all those cases where there are multiple consecutive occurance of NNP; Example: ``\textit{John Jr. Smith is a student}".} \specialspace IS-A=``\textit{is a}" \specialspace O=NN \specialspace ; Ex: ``\textit{John is a student.}"}
\end{enumerate} 

In this case, when the subject or the object is $NNP$ (i.e. the sentence is a ground fact), we use a 4-class NER tool~\footnote{\url{http://nlp.stanford.edu/software/CRF-NER.shtml}} for labeling the corresponding \textit{induced} concepts. Concept induction for ground facts, also called \textit{T-Box induction}, is always entailed whenever an A-Box assertion is made. For example, in the sentence ``\textit{John is a student}", the subject concept labeling will be ``\textit{StudentPerson}". If the NER tool is not able to recognize the NNP, then we plug-in the string ``\textit{Like}"+\footnote{+: String concatenation}``\textit{S}" in place of the string ``\textit{NER(S)}" \footnote{\textit{NER(x)} is a function that outputs the NER category of the named entity \textit{x}}. For example, in the sentence "\textit{Priyansh is a student}" we label the subject concept as ``\textit{StudentLikePriyansh}". In this case, a nominal concept (i.e. $\{Priyansh\}$) will be generated for defining the subject concept. \\

\noindent\textbf{II. Trivial Modifier Rule:} Normally, if a modifier in a IS-A type sentence is \textit{JJ} or \textit{NN} then it modifies either \textit{NN} or \textit{NNP}.  There are two types of modification that can occur: (i) \textit{forward modification} and (ii) \textit{backward modification}. Forward modification is the most commonly observed rule where the modified subject/object concept (i.e.  M \hspace{0.1cm} S/O ) becomes a sub-class of the subject/object. For example, in the sentence: ``\textit{Wild cat is a mammal}" the modifier \textit{Wild} (\textit{JJ} - an adjective here) modifies the subject concept \textit{Cat} (which is a \textit{NN}) and hence, the concept \textit{WildCat} becomes a sub-class of the subject concept \textit{Cat}. \\


\hspace{1.5cm}\begin{minipage}{\textwidth}
\specialpar{\textbf{T-Box Rule} \textit{(Forward Modification)}}
\specialpar{If \specialspace M = NN/JJ \specialspace  S/O = NN \specialspace then} \\
 MS/O $\sqsubseteq$ S/O\\
\end{minipage}

Backward modification is entailed by all such forward modification (if $M=JJ$) in the following way: \\


\hspace{1.5cm}\begin{minipage}{\textwidth}
\specialpar{\textbf{T-Box Rule} \textit{(Backward Modification)}}
\specialpar{If \specialspace M=JJ \specialspace S=NN \specialspace then} \\
MS $\sqsubseteq$ MThing \\
MThing $\equiv \exists $ \textbf{\textit{hasAttribute}}.M\\
\end{minipage}

It is to be noted that $\textbf{\textit{hasAttribute}}$ is a primitive role for \textit{MThing}. In other words, \textit{MThing} is a concept that must have the attribute \textit{M}. In some cases, the backward modification may happen even in the form $M=NN \hspace{7pt} S/O=NN$. For example, in the sentence: ``\textit{House boat is a kind of vessel}." the modifier \textit{House} also gets modified by the subject \textit{Boat} and hence, the subject concept \textit{HouseBoat} is a sub-class of both the concepts \textit{House} and \textit{Boat}. However, such backward modification is not necessarily true for all such cases. For example, in the sentence: ``\textit{Sea plane is an air vehicle}" the subject modifier concept \textit{Sea} is not modified by the concept \textit{Plane} so as to form the concept \textit{SeaPlane} as a sub-class of the concept \textit{Sea} (although the forward modification is still valid). In this example, the same holds true for the modified object (i.e. \textit{AirVehicle}). One approach to capture such semantics is to verify the possibility, using a third party lexicon such as WordNet, whether, say, \textit{SeaPlane} has hyponymy with \textit{Sea}.

 We also observed that there is an exception to the forward modification rule when $M=NN \hspace{7pt} S=NNP$. An example is: ``\textit{King Richard was an English ruler}". In this case a backward modification is required as follows:\\

\hspace{1.5cm}\begin{minipage}{0.8\textwidth}
\textbf{A-Box Rule} \textit{(Backward Modification)}:\\
\textit{If} $[M=NN][S=NNP]$ \textit{Then} 
\\
$(MLikeS/O)(S/O)$ \\
\\ \textbf{T-Box Rule} \textit{(Backward Modification)}:\\
\textit{If} $[M=NN][S/O=NNP]$ \textit{Then} 
\\
$MLikeS/O \sqsubseteq M$ \\
\end{minipage}

%

 Another interesting phenomenon that can be observed for subject/object modification is what we term as \textit{recursive nested forward modification}. In sentences where the subject modification is by a sequence of modifiers such as $M_{1} \hspace{7pt}M_{2} \hspace{7pt} M_{3} \hspace{7pt} S \hspace{7pt} IS-A \hspace{7pt} O$ then a nested structure is assumed as: $M_{1} \hspace{7pt}(M_{2} \hspace{7pt} (M_{3} \hspace{7pt} (S))) \hspace{7pt} IS-A \hspace{7pt} O$. Here ``( )" denotes scope of the modifier. Therefore, the scope of the inner most nested modifier $M_{3}$ is the subject concept \textit{S}. The scope of the modifier $M_{2}$ is the sub-concept $M_{3}S$ formed as a result of the $M_{3}$ modifying \textit{S}. At the same time $M_{2}$ also recursively modifies \textit{S} to form the sub-concept $M_{2}S$. Similarly, $M_{1}$ has the sub-concept $M_{2}M_{3}S$ as scope of modification while in recursion modifies $M_{3}S$ and \textit{S}. The T-Box rule for such recursive nested modification is as follows:\\


\hspace{1.5cm}\begin{minipage}{\textwidth}
\specialpar{\textbf{T-Box Rule} \textit{(Recursive Nested Forward Modification: 3-level nesting)}}
\specialpar{if \specialspace M$_{2}$=JJ/NN \specialspace M$_{1}$=JJ/NN \specialspace S \specialspace then} \\
M$_{1}$S/O $\sqsubseteq$ S/O \\
M$_{2}$M$_{1}$S/O $\sqsubseteq$ M$_{1}$S/O \\
M$_{2}$M$_{1}$S/O $\sqsubseteq$ M$_{2}$S/O \\
\end{minipage}

\noindent\textbf{III. Trivial Subject Quantifier Rule:} We have identified four trivial subject quantifier NSS variation\footnote{Data set - \crdurl}. We hereby describe two of them.

\begin{enumerate}
\item { Q=``\textit{some}" \specialspace S=NNS \footnote{NNS: Plural form of common noun as per Penn Treebank} \specialspace IS-A=``\textit{is a}" \specialspace O=JJ/NN \specialspace ; Ex: ``\textit{Some students are hard-working.}"}
\item {Q=``\textit{the}" \specialspace S=NN \specialspace IS-A=``\textit{is a}" \specialspace O=JJ/NN \specialspace; Ex: ``\textit{The student is hard-working.}"}
\end{enumerate}

%


The first instance is clearly a T-Box statement. The second instance is an A-Box assertion that requires a T-Box induction. For both the cases we need to make a special sub-class of the subject concept. For example, in the sentence: ``\textit{Some students are hard-working}." we need to create a class of students who are hard-working and for that we label the subject concept as \textit{Hard-workingStudent}. The general labeling rule for trivial subject quantification is `\textit{O}'+`\textit{S}'=\textit{OS}. For NSS instance 1, the T-Box axiom is pretty straightforward. For NSS instance 2, we need a corresponding A-Box assertion and their T-Box induction axioms as follows: \\


\hspace{1.5cm}\begin{minipage}{\textwidth}
\specialpar{\textbf{T-Box Rule} \textit{(Trivial Subject Quantifier Rule)}}
\specialpar{if  \specialspace Q$_{1}$=``\textit{some}" \specialspace S \specialspace then} \\
	OS $\sqsubseteq$ S $\sqcap$ O \\
\end{minipage}
An interesting observation in the case of quantifier is that there is a wide range of semantic variation in terms of degree of cardinality for all the possible quantifier lexical variations. For example, the quantifier \textit{not much} usually means low degree of cardinality as compared to the quantifier \textit{quite a lot} which means high degree of cardinality. The quantifiers \textit{some} and \textit{the}, on the other hand, are non-committing in terms of degree of cardinality. This semantic difference of quantifiers cannot be represented in DL adequately, especially when the lexico-syntactic normalizer converts all lexical variations into the neutral quantifier \textit{some}. However, all such normalization (and subsequent characterization and DL translation) maintain a reference to the original sentence which bears the linguistic semantic difference. 

\subsubsection{Non-Trivial DL Translation Rules}
\label{DLNonTrivial}


In this section we will consider five different types of non-trivial non-negative IS-A sentences with their corresponding DL translation. Five different types are (1) similarity sentences (2) subject hypernymy-holonymy ambiguity resolution rule (3) special subject quantifier based sentences (4) sentences with different tense and (5) sentences with modality.\\


\noindent\textbf{I. Non-Trivial Similarity Rule:} The IS-A variation \textit{like} and its variants  pose a very interesting problem since its linguistic semantics is not exactly same as either \textit{is a} or \textit{same as}. For example, a sentence such as ``\textit{Tangerine is like an orange}" does not imply that the concept \textit{Tangerine} is either equivalent to or sub-class of the concept \textit{Orange}. What it means is that the subject concept \textit{Tangerine} and the object concept \textit{Orange} share some common characteristics and can be grouped under one concept representing the commonality. Therefore the translation rules are:\\


\hspace{1.5cm}\begin{minipage}{\textwidth}
\specialpar{\textbf{T-Box Rule} \textit{(Subject-Object Similarity)}}
\specialpar{If \specialspace S \specialspace IS-A=``\textit{is like a}" \specialspace O \specialspace then }\\
S $\sqsubseteq$ SOLike \\
O $\sqsubseteq$ SOLike\\
SOLike $\equiv$ SLike $\sqcap$ OLike\\
\end{minipage}

Here the common parent concept \textit{SOLike} is an intersection of the primitive concepts SLike and OLike.\\

\noindent\textbf{II. Subject Hypernymy-Holonymy Ambiguity Resolution Rule:} For the NSS instance $S=NN \specialspace IS-A=``includes"$ \specialspace $O=NN$ there is an innate ambiguity regarding whether a subject hypernymy rule is to be invoked or whether a subject holonymy rule is to be triggered. For example, in the NSS instance $S=``University" \specialspace IS-A=``includes" \specialspace  O=``faculty"$ the lexicon \textit{includes} entails subject holonymy since \textit{Faculty} has a partitive relation with \textit{University}. However, in other cases such as the NSS instance $S=``\textit{Wild cat}" \specialspace  IS-A=``\textit{includes}" \specialspace  O=``\textit{bob cat}"$, the lexicon \textit{includes} entails subject hypernymy. To resolve this ambiguity we verify the assumed hypernymy in WordNet and if no support is found then we accept the semantics to be subject holonymy. In case it is holonymic in nature then the sentence, strictly speaking, is not an IS-A type sentence. The corresponding T-Box rules will be:\\


\hspace{1.5cm}\begin{minipage}{\textwidth}
\specialpar{\textbf{T-Box Rule} \textit{(Subject Hypernymy / Subject Holonymy)}}
\specialpar{if \specialspace  S=NN \specialspace IS-A=``\textit{includes}" \specialspace O=NN \specialspace \&\& \specialspace If \specialspace ``\textit{includes}" $\rightarrow$ \textit{Hypernymy} \specialspace then}\\
O $\sqsubseteq$ S
\specialpar{else \specialspace (Subject Holonymy)}\\
S $\sqsubseteq \exists$ \textbf{\textit{includeO}}$^{+}$.O\\
\textit{\textbf{includeO}}$^{+} \sqsubseteq$ \textit{\textbf{include}}$^{+}$ \\
\end{minipage}

Here $\textit{\textbf{includeO}}^{+}$ represents the transitive holonymy relation that the subject has with the object. All such roles are sub-properties of the primitive role $\textit{\textbf{include}}^{+}$. 

%

To support ontology evolution, roles are appended with filler to form primitive roles. For example if later on another NSS instance, say  $ S=``University" \specialspace IS-A=``includes" \specialspace O=``department"$, is encountered by the system for DL translation then to expand the earlier definition of \textit{University} we need to use a new role $\textit{\textbf{includeDepartment}}^{+}$ for the filler \textit{Department}. \\

\hspace{1.5cm}\begin{minipage}{0.8\textwidth}
$University \sqsubseteq (\exists \ \textit{\textbf{includeFaculty}}^{+}. \ Faculty \sqcap \exists \ \textit{\textbf{includeDepartment}}^{+}. \ Department)$ \\
\end{minipage}

Use of $\textit{\textbf{include}}^{+}$ instead $\textit{\textbf{includeO}}^{+}$ would make the representation not true both epistemologically and linguistically as (i) \textit{Faculty} and \textit{Department} must have intersection and (ii) the filler concept is this intersection.\footnote{$University \sqsubseteq (\exists \textit{\textbf{include}}^{+}. \ Faculty \sqcap \exists \textit{\textbf{include}}^{+}. \ Department) \equiv \exists \textit{\textbf{include}}^{+}. \ (Faculty \sqcap Department) \equiv \ \perp$ (i.e. epistemologically)}.\\
  
%
%
%
%
%
\noindent\textbf{III. Special Subject Quantifier Rule:}
We have identified a total of 15 basic NSS instances for non trivial NSS instances\footnote{Data set - \crdurl}. We discuss two of them as follows:


\begin{enumerate}
\item{Q=``\textit{At least}" \ CD \ ``\textit{of the}" \specialspace S \specialspace IS-A \specialspace O \specialspace ; Ex: ``\textit{At least one of the students is hard-working.}"}
\item{Q=``Only" \specialspace S \specialspace IS-A \specialspace O \specialspace ; Ex: ``\textit{Only John is a musician.}"}
\end{enumerate}

The scope of quantifiers in such NSS instances is over determined subjects/objects (e.g.: ``\textit{the students ...}", ``\textit{John, Joe, and Mary ...}", etc.). This enforces either a qualititative restriction (e.g.: ``\textit{at least some of the students ...}") or quantitative restriction (e.g.: ``\textit{at least five of the students ...}") on the subject/object. In the first instance the following rule holds:

\hspace{1.5cm}\begin{minipage}[t]{0.84\textwidth}
\specialpar{\textbf{T-Box Rule} \textit{(Q=``\textit{At least}" \ CD \ ``\textit{of the}" \specialspace S=NNS\specialspace IS-A \specialspace O=NN)}}\\
$OS_{N}^{CD_{Min}} \equiv S_N \sqcap O \sqcap \exists \ \textbf{\textit{belongsTo}}. \ \ (\exists \ \textit{\textbf{hasCardinality}}. \ (\textbf{\textit{Cardinality}} \sqcap \exists \ \textbf{\textit{xsd:minInclusive}}. \{CD\}))$ \\
$S_N \sqsubseteq S \sqcap \exists \ \textbf{\textit{belongsTo}}. \ \ (\exists \ \textit{\textbf{hasCardinality}}. \ (\textbf{\textit{Cardinality}} \sqcap \exists \ \textbf{\textit{xsd:minExclusive}}. \{CD\}))$\\
$\{CD\} \sqsubseteq \textbf{\textit{xsd:Integer}}$\\
\end{minipage}

In this representation the concepts and roles in bold faces are primitive and have well-defined set-theoretic semantics. The subject is appended with subscript ``\textit{N}", where \textit{N} denotes the \textit{N}-th unique \textit{group-instance} of the subject. A group-instance is a determined subject that denotes a particular group of instances of that subject. \textit{N} is incremented when the determined subject cannot be co-referenced with any previously recorded group-instance of the subject. In a similar manner, 
the quantifier ``\textit{only}" imposes an exclusive inclusion axiom on the subject/object and hence, explicitly entails no other inclusion possibility. For example, in the sentence: ``\textit{Only John is a musician}", the subject \textit{John} is an instance of the object \textit{musician} exclusively (i.e. nobody else is). \\

If $X$ is not an object and is concept present in $\Delta$, then corresponding translation rule is as follows:


\hspace{1.5cm}\begin{minipage}[t]{0.8\textwidth}
\specialpar{\textbf{T-Box Rule} \textit{(Q=``Only" \specialspace S=NN/NNS/VBG \specialspace IS-A \specialspace O)}}\\
$\forall X \ne O; X \in \Delta,$
\specialpar{if $(S \sqcup X \neq X)$ then}\\
S $\sqsupseteq$ O\\
(X $\sqcap$ O) $\equiv \bot$
\specialpar{else}\\
O $\sqsubseteq$ S $\sqsubseteq$ X\\
\end{minipage}

The A-Box representation of the above rules (i.e. when $S=NNP$) only holds true within the context boundary of the ground assertion. For example, in the sentence ``\textit{Only John is a musician}", we cannot say, from a epistemic point of view, that the class of \textit{student} only contains \textit{John}. However, this is true in the context in which the statement is asserted. To distinguish contexts we label the object with a count $N$ and co-reference this object as and when possible. Hence, in this example the label of the object will be $Musician{N_O}$\footnote{Context identification and corresponding object labeling demands considerable research attention. However, the current DLOL system is completely context-free in this sense.}.\\ 

\noindent\textbf{IV. Tense Rule:} Another very important cause of non-triviality in factual sentences is tense modality. For IS-A type sentences the tense variation of the IS-A lexicon can be either a referral to the past or that to the future. Tense representation is not an easy problem in knowledge discovery research. We have identified a total of six basic NSS instances for Tense Axioms\footnote{Data set - \crdurl} of which three are listed below. 


\begin{enumerate}
\item{ S \specialspace IS-A=VBD \specialspace O \specialspace ; Ex: ``\textit{John was a teacher.}"}
\item{S \specialspace M$_1$=CD \specialspace M$_2$=NNS\_UNIT\_TIME \specialspace M$_3$=``\textit{ago}" \specialspace IS-A=VBD \specialspace O  \specialspace ; Ex: ``\textit{John was a teacher three years ago.}"}
\item{S \specialspace M$_1$=``\textit{for}" \specialspace M$_2$=CD \specialspace M$_3$=NNS\_UNIT\_TIME \specialspace IS-A=VBD \specialspace O \specialspace ; Ex: ``\textit{John was a teacher for three years.}"}
\end{enumerate}

In order to represent all the above NSS instances in DL we introduce three new primitive concepts borrowed from OWL Time\footnote{\url{https://www.w3.org/TR/owl-time/}}: (i) \textbf{\textit{ProperInterval}}, (ii) \textbf{\textit{Instant}}, and (iii) \textbf{\textit{DurationDescription}}. \textbf{\textit{ProperInterval}} denotes a class of time intervals while \textbf{\textit{Instant}} denotes a class of timestamps. \textbf{\textit{DurationDescription}} is a class of different time units such as \textit{years}, \textit{months}, etc. We also introduce a primitive role called \textbf{\textit{isTrueFor}} that defines the ontological validity of any concept with respect to a particular time interval. The representations for instances one through three is given as follows: \\


\hspace{1.5cm}\begin{minipage}[t]{0.8\textwidth}
\specialpar{\textbf{T-Box Rule} (S \specialspace IS-A=VBD \specialspace O )}\\
\textit{Let} $t_{pr}$ \textit{represents the current time;} $t_{pr}' < t_{pr}$\\
$S \sqsubseteq O \sqcap \exists \ \textbf{\textit{isTrueFor}}. \ (\textbf{\textit{ProperInterval}} \sqcap \exists \ \textbf{\textit{hasEnd}}. \ (\textbf{\textit{Instant}} \sqcap \exists \ \textbf{\textit{inDateTime}}. \ \{t_{pr}'\}))$\\
$\{t_{pr}'\} \sqsubseteq \textbf{\textit{Instant}}$\\ 
\end{minipage}

Sentences for second NSS instance are different from the sentences for NSS instance as they quantify the time in past. \\


\hspace{1.5cm}\begin{minipage}{0.8\textwidth}
\specialpar{\textbf{T-Box Rule} (S \specialspace M$_1$=CD \specialspace M$_2$=NNS\_OWLTIME\footnotemark \specialspace M$_3$=``\textit{ago}" \specialspace IS-A=VBD \specialspace O)}\\
\textit{Let} $t_{pr}$ \textit{represents the current time;}\\
$S \sqsubseteq O \sqcap \exists \ \textbf{\textit{isTrueFor}}. \ (\textbf{\textit{ProperInterval}} \sqcap \exists \ \textit{\textit{\textbf{intervalMeets}}}. \ (\textbf{\textit{ProperInterval}} \sqcap \exists \ \textbf{\textit{hasEnd}}. \ (\textbf{\textit{Instant}} \sqcap \exists \ \textbf{\textit{inDateTime}}. \ \{t_{pr}\}) \sqcap \exists \ \textbf{\textit{hasDurationDescription}}. \ (M_2 \sqcap \textbf{\textit{DurationDescription}} \sqcap \exists \ OWLTIME. \ \{M_1\})))$\\
$\{t_{pr}\} \sqsubseteq \textbf{\textit{Instant}}$\\
$\{M_1\} \sqsubseteq \textbf{\textit{Decimal}}$\\ 
\end{minipage}

\footnotetext{OWLTIME is a special tag variable annotated by the special annotator that can have value from \{\textit{years, months, weeks, days, hours, minutes, seconds}.\}}

Sentence for third NSS instances is different from the sentences for first NSS instance and second NSS instance as they specify the duration. \\


\hspace{1.5cm}\begin{minipage}{0.8\textwidth}
\specialpar{\textbf{T-Box Rule} \textit{(S \specialspace M$_1$=``\textit{for}" M$_2$=CD \specialspace M$_3$=NNS\_UNIT\_TIME \specialspace IS-A=VBD \specialspace O)}}\\
\textit{Let} $t_{pr}$ \textit{represents the current time;} $t_{pr}' < t_{pr}$\\
$S \sqsubseteq O \sqcap \exists \ \textbf{\textit{isTrueFor}}. \ (\textbf{\textit{ProperInterval}} \sqcap \exists \ \textbf{\textit{hasEnd}}. \ (\textbf{\textit{Instant}} \sqcap \exists \ \textbf{\textit{inDateTime}}. \ \{t_{pr}'\}) \sqcap \exists \ \textit{\textit{\textbf{hasDurationDescription}}}. \ (M_3 \sqcap \textbf{\textit{DurationDescription}} \sqcap \exists \ OWLTIME. \ \{M_2\}))$\\
$\{t_{pr}'\} \sqsubseteq \textbf{\textit{Instant}}$\\
$\{M_2\} \sqsubseteq \textbf{\textit{Decimal}}$\\
\end{minipage}


\noindent\textbf{V. Modal Rule:} Apart from tense modality there is yet another kind of modality, called \textit{epistemic modality}, whose formal representation has drawn significant research attention as well. The peculiarity of epistemic modality is that it entails probabilistic concept membership/inclusion. The following are two out of seven basic NSS instances\footnote{Data set - \crdurl} that require modal axiomatic representation:

\begin{enumerate}
\item{S \specialspace IS-A=``\textit{may be}" \specialspace O \specialspace; Ex: ``\textit{John may be a good swimmer.}"}
\item{S \specialspace IS-A=``\textit{can become}" \specialspace O \specialspace; Ex: ``\textit{John can become a diligent researcher.}"}
\end{enumerate}

In order to represent the modality we define a notion of using probability function $Pr( \cdot , \cdot )$ where the first argument is to consider the membership of the instance of the subject and second argument is to consider the time at which the statement was made. In order to capture the condition for probability to be greater than zero, boolean function $greaterThan( \cdot , \cdot )$ is used. Representation for the first NSS instance is as follows :\\


\hspace{1.5cm}\begin{minipage}[t]{0.8\textwidth}
\specialpar{\textbf{T-Box Rule} \textit{(S \specialspace IS-A=``\textit{may be}" \specialspace O)}}\\
$\textbf{\textit{mayBe}}^I = \{(x,y) \in \Delta^I \times \Delta^I \ | \ \exists (S,O) \ \exists t_{pr}; \ S^I \subseteq \Delta^I \wedge O^I \subseteq \Delta^I \wedge x \in S^I \wedge y \in O^I \rightarrow greaterThan(Pr(O(x), t_{pr}'), 0)\}$\\
$S \equiv \exists \ \textbf{\textit{mayBe}}.O$\\
\end{minipage}

Sentence to belong second NSS instance introduce the modality about the ability of the subject being related to object and is represented in a manner similar to first NSS instance but with the consideration of the threshold. \\


\hspace{1.5cm}\begin{minipage}[t]{0.8\textwidth}
\specialpar{\textbf{T-Box Rule} \textit{(S \specialspace IS-A=``\textit{can become}" \specialspace O)}}\\
$\textbf{\textit{canBecome}}^I = \{(x,y) \in \Delta^I \times \Delta^I \ | \ \exists (S,O) \ \exists (t_{pr},t_{pr}'); \ S^I \subseteq \Delta^I \wedge O^I \subseteq \Delta^I \wedge x \in S^I \wedge y \in O^I \rightarrow greaterThan(Pr(O(x), t_{pr}'), 0) \wedge greaterThan(t_{pr}',t_{pr}) \wedge equals(Pr(O(x),t_{pr}), 0))\}$\\
$S \equiv \exists \ \textbf{\textit{canBecome}}.O$\\
\end{minipage}

\subsection{DLT Algorithm}
\label{DLT}
DL Translator (DLT) algorithm converts an NSS instance into a corresponding DL expression. The object phrase is first extracted from the NSS Instance. If the object contains  quantifiers or modifiers, corresponding modifier or quantifier rule is invoked. In this way, the DL expression of the object is formed. Then the IS-A phrase is extracted and corresponding rules of hyponymy, hypernymy, and/or modal IS-A phrases are invoked, thus, generating the DL expression for it. Finally, the subject phrase is extracted, and its modifiers and quantifiers are detected in the same manner as in the case of object phrase. The corresponding translation rules are invoked to generate a DL expression. Finally, a global DL expression is generated by combining the DL expressions of the subject, object and IS-A phrase. The NSS instance parsing precedence followed while calling base rules are as follows (in descending order):
\begin{enumerate}
	\item{\textbf{Object Phrase DL Sub-expression Generation}}
	\begin{enumerate}
		\item Object Tag (POS Tag \& Special Annotation)
		\item Object Modifier
		\item Object Quantifier
	\end{enumerate}
	\item {\textbf{IS-A Phrase DL Sub-expression Generation}}
	\begin{enumerate}
		\item IS-A type (hyponymy, hypernymy, modal, etc.)
	\end{enumerate}
	\item{\textbf{Subject Phrase DL Sub-expession Generation}}
	\begin{enumerate}
		\item Subject Tag (POS Tag \& Special Annotation)
		\item Subject Modifier
		\item Subject Quantifier 
	\end{enumerate}
	
\end{enumerate}
It combines the DL sub-expressions formed per call. The resultant DL expression is stored into the knowledge base. The algorithm pseudo-code is as follows:

\begin{algorithm*}[h]
	\KwData{$NSS \ Instance$}
	\KwResult{DL Translator Algorithm}
	\Begin{
		$NSS\_type \longleftarrow GET\_NSS\_TYPE(\textit{NSS\_Instance})$\;
		$object\_DL \longleftarrow CALL\_OBJECT\_DL\_RULE(NSS\_Type, \ GET\_OBJECT(\textit{NSS\_Instance}))$\;
		$IS-A\_DL \longleftarrow CALL\_IS-A\_DL\_RULE(NSS\_Type, \ GET\_IS-A(\textit{NSS\_Instance}))$\;
		$subject\_DL \longleftarrow CALL\_SUBJECT\_DL\_RULE(NSS\_Type, \ GET\_SUBJECT(\textit{NSS\_Instance}))$\;
		$GENERATE\_DL\_EXPRESSION(object\_DL, \ IS-A\_DL, \ subject\_DL)$\;
	}
\caption{DLT Algorithm
\label{DLT}}
\end{algorithm*}

It is to be noted that the \textit{DL Translator} internally calls a labeling algorithm for creating labels for derived concepts. The labeling algorithm takes into consideration: (i) subject or object POS tag, (ii) forward and backward modification, (iii) whether subject or object is quantified with determiners.
	
\section{Evaluation}
\label{evaluation}
\subsection{Evaluation Goals and Metrics}
We evaluate \dlol \ from two different perspectives: (i) syntactic robustness of NSS template-fitting algorithm, and (ii) semantic accuracy of DLT algorithm using gold-standard benchmark evaluation (as described in \cite{85dellschaft2006perform}). We describe each of the evaluation goals below:

\textbf{Goal I. Evaluation of Syntactic Robustness:} Syntactic robustness of NSS can be understood by analyzing the soundness and completeness of the NSS template-fitting algorithm. By soundness, we mean that there should not be any factual IS-A type English sentence which is ``\textit{incorrectly fitted}" into the template. Here, correct-fitting implies that there should not be any mismatch between the POS-tag of a linguistic constituent and its corresponding NSS cell. By completeness, we mean that there should not be any valid factual IS-A type English sentence that is not accepted by the algorithm, either fully or partially. To evaluate template-fitting accuracy, we came up with two measures: (i) \textit{Characterization Precision}, and (ii) \textit{Characterization Recall}.

\begin{mydef}
\textbf{Characterization Precision (CP)}: It is the ratio of the number of correctly fitted sentences ($N_{CF}$) to the total number of fitted sentences ($N_F$).
\end{mydef}

\begin{mydef}
\textbf{Characterization Recall (CR)}: It is the ratio of correctly fitted sentences ($N_{CF}$) to the total number of sentences ($N$) in the test IS-A corpus.
\end{mydef}

\textbf{Goal II. OL Accuracy:}
In order to evaluate the accuracy of OL in terms of \dlol's term extraction capability, we use two commonly adopted measures \cite{85dellschaft2006perform}: (i) \textit{Lexical Precision} (\textit{LP}), and (ii) \textit{Lexical Recall} (\textit{LR})\footnote{Formal definitions can be found in \cite{4wong2012ontology}.}. LP/LR indirectly measures the false positive and false negative during a concept-extraction process. On the other hand, to understand the structural accuracy of the learned ontology (i.e. the semantic accuracy of DLT algorithm), we also require a measure that can compute labeled-graph similarity between learned and engineered ontologies, since, in general, taxonomies are graphs. We have devised a very effective and easy measure, called  \textit{Instance-based Inference Measure} (\textit{IIM}), for this purpose.


 The idea behind \textit{IIM} is to populate every concept (primitive and defined) in a given ontology with one unique arbitrary instance and then run a DL reasoner (in our case FACT++ \cite{82tsarkov2006fact++}) to classify all the concepts. This leads to a set of \textit{Inferred Instances} (called \textit{II}) within each class. For example, consider the ontology shown in figure \ref{fig:iim}(a). Each concept, AgileAthlete, Athlete, Student, StudiousStudent and Owl:Thing, are initiated with an unique arbitrary instances X$_1$, X$_2$, X$_3$, X$_4$ and X$_5$ (figure \ref{fig:iim}(b)). Once the ontology is classified, instances are inferred such that the inferred instances in each concept reflect the topology of the ontology. Here, X$_4$ and X$_1$ indicates that AgileAthlete is a subclass of Athlete. The following theorem provides the soundness of the measure:

	\begin{figure*}[!t]	
		\centering
		\subfigure[Snapshot of the ontology before aribtary instances are added to the class.]{\includegraphics[scale=0.38]{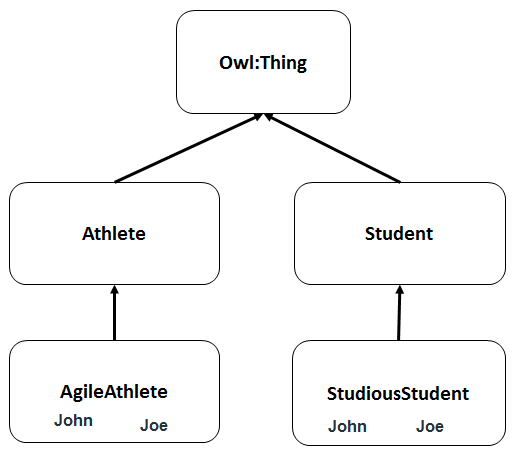}}\quad
		\subfigure[Snapshot of the knowledge Base after unique arbitary instances are added in each class.] {\includegraphics[scale=0.38]{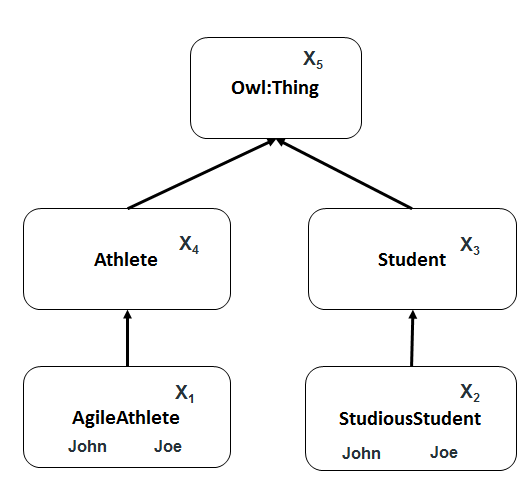}}\quad
		\subfigure[Shapshot of the Knoweldge Base after instances are inferred in each class based on the topology of the Knowledge Base]{\includegraphics[scale=0.38]{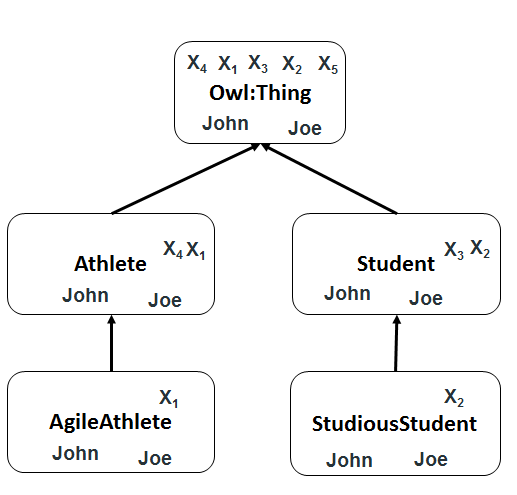}}
		\caption{Example to demonstrate working of Instance-based Inference Measure (IIM). Here a square box indicates a concept and directed edges indicate relationship from a subclass to a superclass.}
		\label{fig:iim}
	\end{figure*}

\begin{theorem}
If two taxonomies are isomorphic and if the pre-classified concepts of the individual taxonomies, having same labels, are instantiated equivalently, then, after the taxonomies are classified, the same-labeled concepts from each of the taxonomies will have equivalent inferred instances.
\end{theorem}
We formally define the IIM accuracy measures as follows:
\begin{mydef}
\textbf{\textit{IIM-Precision} (\textit{IIM-P})}: \textit{IIM-P} of a given learned ontology $\mathcal{\bigtriangleup}_{l}$ with respect to a gold standard ontology $\mathcal{\bigtriangleup}_{GS}$ is defined as:\\
\end{mydef}

\textit{IIM-P}$(\mathcal{\bigtriangleup}_{\textit{l}}, \mathcal{\bigtriangleup}_{\textit{GS}}) = \frac{\sum_{i \in \textbf{CC}_{\mathcal{\bigtriangleup}_{\textit{l}}, \ \mathcal{\bigtriangleup}_{GS}}} \left|II^i_{\mathcal{\bigtriangleup}_{\textit{l}}} \cap II^i_{\mathcal{\bigtriangleup}_{GS}}\right|}{\sum_{i \in \textbf{C}_{\mathcal{\bigtriangleup}_{\textit{l}}}}\left|II^i_{\mathcal{\bigtriangleup}_{\textit{l}}}\right|}$\\ where, \\
$CC_{\mathcal{\bigtriangleup}_{\textit{l}}, \mathcal{\bigtriangleup}_{GS}}$: Set of common concepts in $\mathcal{\bigtriangleup}_{l}$ and $\mathcal{\bigtriangleup}_{GS}$. \\ $C_{\mathcal{\bigtriangleup}_{\textit{l}}}$: Set of concepts in $\mathcal{\bigtriangleup}_{\textit{l}}$.\\

	\begin{figure*}[!t]	
		\centering
		\subfigure[Base (or gold standard) ontology engineered by an expert.]{\includegraphics[scale=0.4]{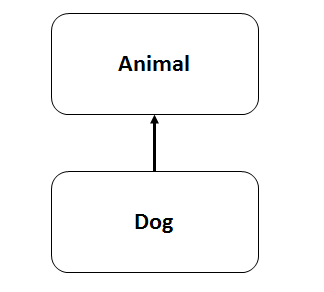}}\hfill
		\subfigure[Variant of the base (or gold standard) ontology learned by a system which misses a relation between Animal and Dog] {\includegraphics[scale=0.4]{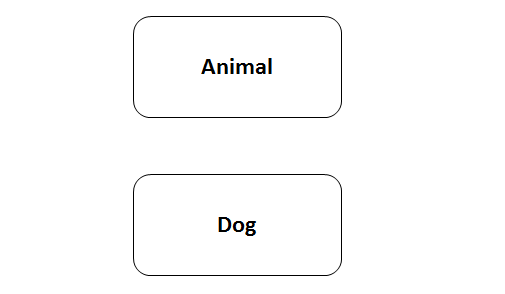}}\hfill
		\subfigure[Variant of the base ontology which contains all the relations of base ontology but happens to introduce a new relation between Animal and Cat.]{\includegraphics[scale=0.4]{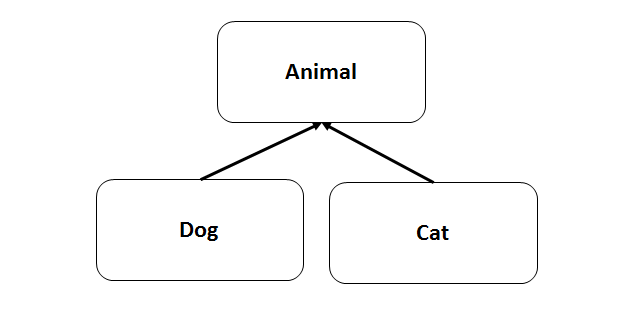}}
		\caption{Example to demonstrate working of IIM-P. Here square box represents a concept and a directed edge represents relationship from a subclass to a superclass.}
		\label{fig:iimp}
	\end{figure*}

IIM-P helps us to understand the noise associated with the overall graph similarity that has been acquired  by an OL tool with respect to the base human-engineering ontology. Noise is caused by additional child concepts in the learned ontology. It should be noted that noise can sometimes be a false alarm since additional child concepts may be an ontologically valid (although possibly unimportant from a pragmatic point of view) identification by an OL tool. Hence \textit{IIM-P} cannot be a first category evaluation parameter. 

We explain this with the help of figures given above. In figure \ref{fig:iimp}, the learned ontology \ref{fig:iimp}(b) could not detect the IS-A relation between Animal and Dog, thereby rendering the learned ontology as incorrect. However IIM-P measure would consider learned ontology to be precise (i.e no noise). This is due to the fact that the learned ontology contains no \textit{extra} relations. While in the case of figure \ref{fig:iimp}(c), the learned ontology successfully contains all the IS-A relations of the base ontology, as well as one extra relation. Presence of an extra relation will be successfully detected by IIM-P implying that the learned ontology is not precise, irrespective of it being correct or not.

\begin{mydef}
\textbf{\textit{IIM-Recall} (\textit{IIM-R})}: \textit{IIM-R} of a given learned ontology $\mathcal{\bigtriangleup}_l$ with respect to a base ontology $\mathcal{\bigtriangleup}_{\textit{GS}}$ is defined as:\\
\end{mydef}
\textit{IIM-R}$(\mathcal{\bigtriangleup}_l, \mathcal{\bigtriangleup}_{\textit{GS}}) = \frac{\sum_{i \in CC_{\mathcal{\bigtriangleup}_l, \mathcal{\bigtriangleup}_{GS}}} \left|II^i_{\mathcal{\bigtriangleup}_l} \cap II^i_{\mathcal{\bigtriangleup}_{\textit{GS}}}\right|}{\sum_{i \in C_{\mathcal{\bigtriangleup}_{\textit{GS}}}}\left|II^i_{\mathcal{\bigtriangleup}_{\textit{GS}}}\right|}$\\ where, \\
$CC_{\mathcal{\bigtriangleup}_{\textit{l}}, \mathcal{\bigtriangleup}_{GS}}$: Set of common concepts in $\mathcal{\bigtriangleup}_{l}$ and $\mathcal{\bigtriangleup}_{GS}$. \\ $C_{\mathcal{\bigtriangleup}_{\textit{GS}}}$: Set of concepts in $\mathcal{\bigtriangleup}_{\textit{GS}}$.\\

\textit{IIM-R} helps us to understand the incompleteness associated with the overall graph similarity. Incompleteness in learned ontology is caused due to missing valid child concepts. \textit{IIM-R} is a better accuracy measure than \textit{IIM-P} since it can never raise a false alarm (assuming a well-engineered gold-standard ontology). In some cases it is more useful to analyze IIM accuracy on the topology that is shared by the two taxonomies (rather than the entire topology). In other words, an OL process may be deemed sufficiently efficient if the common concepts extracted have similar inter-hyponymic structure. Hence, we introduce two additional variations of IIM: \textit{IIM-OP} and \textit{IIM-OR} as follows:
\begin{mydef}
\textbf{\textit{IIM-OverlapPrecision} (\textit{IIM-OP})}: \textit{IIM-OP} of a given learned ontology $\mathcal{\bigtriangleup}_l$ with respect to a gold-standard ontology $\mathcal{\bigtriangleup}_{GS}$ is defined as:\\

\textit{IIM-OP}$(\mathcal{\bigtriangleup}_l, \mathcal{\bigtriangleup}_{GS}) = \frac{\sum_{i \in CC_{\mathcal{\bigtriangleup}_l, \mathcal{\bigtriangleup}_{GS}}} \left|II^i_{\mathcal{\bigtriangleup}_l} \cap II^i_{\mathcal{\bigtriangleup}_{GS}}\right|}{\sum_{i \in CC_{\mathcal{\bigtriangleup}_l, \mathcal{\bigtriangleup}_{GS}}}\left|II^i_{\mathcal{\bigtriangleup}_l}\right|}$\\ where, \\
$CC_{\mathcal{\bigtriangleup}_{\textit{l}}, \mathcal{\bigtriangleup}_{GS}}$: Set of common concepts in $\mathcal{\bigtriangleup}_{l}$ and $\mathcal{\bigtriangleup}_{GS}$. \\
\end{mydef}

\begin{mydef}
\textbf{IIM-OverlapRecall (\textit{IIM-OR})}: \textit{IIM-OR} of a given ontology $\mathcal{\bigtriangleup}_l$ with respect to a gold-standard ontology $\mathcal{\bigtriangleup}_{GS}$ is defined as:\\

\textit{IIM-OR}$(\mathcal{\bigtriangleup}_l, \mathcal{\bigtriangleup}_{GS}) = \frac{\sum_{i \in CC_{\mathcal{\bigtriangleup}_l, \mathcal{\bigtriangleup}_{GS}}} \left|II^i_{\mathcal{\bigtriangleup}_l} \cap II^i_{\mathcal{\bigtriangleup}_{GS}}\right|}{\sum_{i \in CC_{\mathcal{\bigtriangleup}_l, \mathcal{\bigtriangleup}_{GS}}}\left|II^i_{\mathcal{\bigtriangleup}_{GS}}\right|}$\\ where, \\
$CC_{\mathcal{\bigtriangleup}_{\textit{l}}, \mathcal{\bigtriangleup}_{GS}}$: Set of common concepts in $\mathcal{\bigtriangleup}_{l}$ and $\mathcal{\bigtriangleup}_{GS}$. \\
\end{mydef}

As opposed to \textit{IIM-P} and \textit{IIM-R}, \textit{IIM-OP} and \textit{IIM-OR} considers only common concepts present in gold standard ontology and generated ontology. \textit{IIM-OP} helps us to understand the structural noise in the matched lexical space while \textit{IIM-OR} helps us to understand the structural incompleteness in the matched lexical space. We measured OL accuracy also in terms of \textit{degree of semantic preservation}, which is a human-judgment based qualitative measure of analyzing the extent to which the model-theoretic semantics of the DL translation of a natural language sentence is equivalent to the linguistic semantics \footnote{It is to be noted that the formal language is DL in the specific context of Web Ontology Learning. Theoretically, the representation can be any formal language that has a model-theoretic interpretation.}. In order to evaluate semantic preservation, we first manually re-translate every DL definition (and axiom) into its corresponding natural language sentence. As an example, imagine that the sentence ``\textit{Rickshaw is a kind of vehicle that has three wheels.}" gets translated into the DL expression: $Rickshaw \sqsubseteq ThreeWheeledVehicle \sqsubseteq Vehicle  \sqcap  (\ \mbox{=$3$} \  \mbox{\textit{include}}^{+}. \ Wheel)$. Now the expression can be re-translated into the sentence: ``\textit{Rickshaw is a three wheeled vehicle.}". This sentence can be evaluated to be semantically equivalent (in the linguistic sense) to the original sentence. Degree of Semantic Preservation can take three values: (i) \textit{yes}, (ii) \textit{no}, and (iii) \textit{partial}. \textit{Yes} corresponds to complete preservation of the semantics of a given sentence, \textit{no} for total failure, and \textit{partial} if atleast the core semantics of the original sentence is completely captured. We define failed, partial, and complete semantic preservation as follows:
\begin{mydef}
	\textbf{Failed Semantic Preservation}: If the subject-object dependency in a given natural language sentence has been incorrectly identified by the NSS template-fitting algorithm (i.e. CP value is 0 w.r.t the sentence), then the corresponding DL expression is considered to have failed in terms of semantic preservation.
\end{mydef}

\begin{mydef}
\textbf{Partial Semantic Preservation}: If the subject-object dependency in a given natural language sentence has been correctly identified by the NSS template-fitting algorithm (i.e. CP value is 1 w.r.t the sentence), then the corresponding DL expression is considered to hold partial semantic preservation.
\end{mydef}

It is to be noted that in most cases NSS template-fitting algorithm will work even  without any sentence simplification step. As an example, consider  the sentence: ``\textit{It is not true that John happens to be an intelligent student}". Here, the sentence does not get simplified into: ``\textit{John is not an intelligent student}". Then the (correct) NSS instance will be: \specialspace S$_1$=``It" \specialspace R$_1$=``\textit{is not}" \specialspace O$_1$=``\textit{true}" \specialspace Cl= ``\textit{that}" \specialspace S$_2$=``\textit{John}" \specialspace R$_2$=``\textit{happens to be}" \specialspace Q$_2$=``\textit{an}"\specialspace  M$_2$=``\textit{intelligent}" \specialspace O$_2$=``\textit{student}". However, the formal semantic construction during DL translation is going to be erroneous if the simplification does not happen. Also, the redundant phrase ``\textit{It is true that ...}" will add noise to the semantic construction. This leads to partial semantic preservation.

\begin{mydef}
	\textbf{Complete Semantic Preservation}: If the DL expression of a given natural language sentence holds partial semantic preservation, and if model-theoretic semantics of the expression is equivalent to the linguistic reading of the given sentence, then the DL expression is considered to hold complete semantic preservation. 
\end{mydef}


\begin{table}[t]
	\centering
	\caption{Dataset Distribution}
	\smallskip 
	\begin{tabular}{ccccc}
		\hline
		\multicolumn{1}{c}{{\textbf{Dataset}}} &{\textbf{\# Sentences}} & \textbf{\# non IS-A} & \multicolumn{2}{c}{\textbf{\# extracted IS-A ($N$)}} \\
		\hline
		\multicolumn{1}{c}{} &       & \textbf{} & \textbf{Trivial} & \textbf{Non-Trivial} \\
		WCL v. 1.1 & 1777  & 240   & 1537  & 0 \\
		Vehicle & 150   & 0     & 150   & 0 \\
		Virus & 172   & 0     & 171   & 1 \\
		Plant & 638   & 5     & 628   & 5 \\
		\hline
	\end{tabular}
	\label{tab:datasetDistribution}%
\end{table}%

\subsection{Experimental Setup \& Dataset}
\subsubsection{Syntactic Robustness}
For evaluating syntactic robustness, we chose two types of datasets: (i) domain-independent, and (ii) domain-specific. For the first type, we used the WCL v.1.1 (wiki\_good.txt) test corpus \cite{14navigli2010annotated}. This dataset was chosen because: (i) it is a non-negative factual dataset, (ii) it is mostly extracted from Wikipedia covering broad range of topics, (iii) it is mostly definitional, thereby including lot of IS-A type sentences. For the second type of dataset, we used the Vehicle, Virus, and Plant corpus, as extracted from Yahoo! Boss by \cite{49kozareva2010semi}. The details of the dataset statistics is given in Table \ref{tab:datasetDistribution}. We employed two research assistants to create a subset of purely IS-A sentences from these documents and validated the dataset by a professional linguist. CP and CR values were manually computed by them. 


	\begin{table}[t]
		\centering
		\caption{DLOL Characterization Performance}
		\begin{tabular}{ccccccc}
			\hline
			\multicolumn{1}{c}{\textbf{Dataset}} & \textbf{$N$} & \textbf{Fail} & \textbf{$N_{CF}$} & \textbf{POS-tag Error} & \textbf{CP} & \textbf{CR} \\
			\hline
			WCL v. 1.1 & 1537  & 9     & 1528  & 9     & 1     & 0.9941 \\
			Vehicle & 150   & 3     & 147   & 3     & 1     & 0.98 \\
			Virus & 172   & 9     & 163   & 9     & 1     & 0.9477 \\
			Plant & 633   & 2     & 631   & 2     & 1     & 0.9968 \\
			\hline
		\end{tabular}
		\label{tab:dlolCharPerformance}
	\end{table}

\begin{table}[t]
	\label{OL accuracy}
	\centering
	\caption{Topological Characteristics of Engineered/Learned Ontology on WCL IS-A Dataset}
	
	\smallskip 
	\begin{tabular}{cccccrrrr}
		\hline
		\multicolumn{1}{c}{Ontology} & \multicolumn{3}{c}{Concepts(C)} &
		\multicolumn{1}{c}{$N_D/N_{Pr}$ Ratio} &
		\multicolumn{2}{c}{$C_{parent}/C$}\footnotetext{Number of parent concepts per concept} & \multicolumn{2}{c}{$C_{sibling}/C$}\footnotetext{Number of sibling concepts per concept} \\
		
		\multicolumn{1}{c}{} & $N_{T}$& $N_{Pr}$& $N_{D}$& & average & max & average & max \\
		\hline
		Engineer A & 4236  & 4007  & 229   & 0.0571 & \multicolumn{1}{c}{3} & \multicolumn{1}{c}{5} & \multicolumn{1}{c}{14} & \multicolumn{1}{c}{40} \\
		Engineer B & 3134  & 2928  & 207  & 0.0707 & \multicolumn{1}{c}{3} & \multicolumn{1}{c}{6} & \multicolumn{1}{c}{12} & \multicolumn{1}{c}{40} \\
		Engineer C & 3191  & 2893  & 298 & 0.1030  & \multicolumn{1}{c}{3} & \multicolumn{1}{c}{6} & \multicolumn{1}{c}{13} & \multicolumn{1}{c}{38} \\
		DLOL  & 4507  & 4320  & 187 & 0.0432  & \multicolumn{1}{c}{4} & \multicolumn{1}{c}{7} & \multicolumn{1}{c}{22} & \multicolumn{1}{c}{92} \\
		FRED  & 3396  & 1799  & 1597 & 0.8877 & \multicolumn{1}{c}{4} & \multicolumn{1}{c}{6} & \multicolumn{1}{c}{9} & \multicolumn{1}{c}{20} \\
		LExO & 6375  & 1406  & 4969 & 3.5341 & \multicolumn{1}{c}{9} & \multicolumn{1}{c}{28} & \multicolumn{1}{c}{10} & \multicolumn{1}{c}{30} \\
		Text2Onto & 1791  & 1791  & 0  & 0  & \multicolumn{1}{c}{2} & \multicolumn{1}{c}{2} & \multicolumn{1}{c}{11} & \multicolumn{1}{c}{33} \\
		\hline
	\end{tabular}%
	\label{tab:ontologyMatrix}
\end{table}%

\subsubsection{OL Accuracy}		
To evaluate OL accuracy we compared, \dlol~ with three different OL tools/framework: (i) Text2Onto, (ii) FRED, and (iii) LExO. We chose Text2Onto because it represents the widely adopted lexico-syntactic pattern based OL paradigm. FRED was chosen because it represents the paradigm of OL tools that attempt to perform deep linguistic analysis using sophisticated linguistic tools. LExO was chosen because it represents the paradigm of rule-based axiomatic OL through analysis of linguistic pattern structures. It is to be noted that Text2Onto is an OL framework (and not a tool per se). However, by properly choosing the required algorithms (for concept extraction, IS-A and non IS-A relation generation, etc.) we can generate a final OWL ontology from a given text corpus. \textit{FRED} is a proper OL tool\footnote{\url{http://wit.istc.cnr.it/stlab-tools/fred}}. LExO can be seen as an OL tool from an axiom generation perspective (since axioms form a generalized ontology). 
For conducting the comparative evaluation in terms of LP/LR and IIM-P/IIM-R, we used the WCL IS-A dataset that has been described in the previous section. In lieu to standard norm followed in gold-standard OL evaluation, we employed three ontology engineers for designing three independent versions of benchmark ontologies on this dataset. From Table \ref{tab:ontologyMatrix} we can observe that all the three engineers agreed that the number of primitives ($N_{Pr}$) is lot more for the WCL dataset as compared to derived concepts ($N_{D}$), with respect to total number of concepts ($N_T$). In other words, the engineered ontologies are mostly axiomatic than definitional. Although WCL IS-A dataset helps to analyze performance of OL tools on IS-A type sentences, yet it has to be noted that the dataset does not comprehensively cover all the IS-A types (both trivial and non-trivial) that have been identified in this paper. Hence, we propose a community test dataset\footnote{Dataset available at \crdurl} in order to measure the \textit{degree of semantic preservation} (as described in the previous section). The dataset comprises of 15 distinct types of trivial IS-A sentences and 56 distinct types of non-trivial sentences.

Since the objective of this dataset is to evaluate general purpose ontology learning systems, the ontology thus created was not guided by some domain specific ontology engineering principles.
The design choices taken by the ontology engineer thus were primarily guided by their interpretation of the original sentence, their understanding of the \textit{taxonomy}, i.e. their cognitive interpretation of the taxonomical relationships and entities present in the sentences.

	\begin{table*}[t]
		\centering
		\caption{OL Accuracy: Comparative Analysis of \dlol~ on WCL IS-A Dataset. Best performance is marked in bold}
		\smallskip 
		\resizebox{\linewidth}{!}{%
			\begin{tabular}{|*{13}{c|}}
				\hline
				\textbf{Measure} & \multicolumn{3}{c|}{\textbf{DLOL  ($N_T=$ 4507)}} & \multicolumn{3}{c|}{\textbf{FRED ($N_T=$ 3396)}} & \multicolumn{3}{c|}{\textbf{LExO ($N_T=$ 6375)}} & \multicolumn{3}{c|}{\textbf{Text2Onto ($N_T=$ 1791)}}\\
				\cline{2-13}
				& \textbf{Ver. A} &  \textbf{Ver. B} &  \textbf{Ver. C} & \textbf{Ver. A} & \textbf{Ver. B} & \textbf{Ver. C}  & \textbf{Ver. A} &  \textbf{Ver. B} &  \textbf{Ver. C} & \textbf{Ver. A} & \textbf{Ver. B} & \textbf{Ver. C}\\
				\hline
				\textbf{LP}    & 0.6381 & 0.5236 & 0.5283 & 0.6346 & 0.5980 & 0.6054 & 0.1810 & 0.1856 & 0.1867 & \textbf{0.7370} & \textbf{0.7320} & \textbf{0.7381} \\
				\hline
				\textbf{LR}    & \textbf{0.6789} & \textbf{0.7530} & \textbf{0.7462} & 0.5087 & 0.6480 & 0.6443 & 0.2723 & 0.3774 & 0.3713 & 0.3116 & 0.4183 & 0.4143 \\
				\hline
				\textbf{IIM-P} & 0.3517 & 0.3106 & 0.3098 & 0.0773 & 0.0677 & 0.0682 & 0.1809 & 0.1851 & 0.1859 & \textbf{0.4924} & \textbf{0.5324} & \textbf{0.5341} \\
				\hline
				\textbf{IIM-R} & \textbf{0.4660} & \textbf{0.5286} & \textbf{0.5222} & 0.2528 & 0.2845 & 0.2835 & 0.0952 & 0.1249 & 0.1263 & 0.1440 & 0.2000 & 0.1988 \\
				\hline
				\textbf{IIM-OP} & 0.5523 & 0.5589 & 0.5550 & 0.1057 & 0.1077 & 0.1080 & \textbf{1.0000} & \textbf{1.0000} & \textbf{1.0000} & 0.5580 & 0.6630 & 0.6618 \\
				\hline
				\textbf{IIM-OR} & \textbf{0.6042} & \textbf{0.6421} & \textbf{0.6430} & 0.3843 & 0.4027 & 0.4045 & 0.2869 & 0.2918 & 0.2987 & 0.3005 & 0.3874 & 0.3881 \\
				\hline
			\end{tabular}}%
			\label{OLAccuracy} %
		\end{table*}%
	
\subsection{Results}
\subsubsection{Syntactic Robustness: CP \ / \ CR} We observed 100\% average CP and 98\% average CR when we tested the NSS template-fitting algorithm on the chosen datasets (Table \ref{tab:dlolCharPerformance}). This empirically shows that NSS is robust on both generic and domain-specific datasets. The perfect CP value is because the NSS template-fitting algorithm did not find any ambiguity in identifying the subject-object dependencies. This can be attributed to the fact that most sentences in the datasets are trivial IS-A type. In comparison to precision we observed a slightly lower recall. This was because of inaccurate POS-tagging of lexemes, which may lead to: (a) inaccurate sentence simplification, and/or (b) invalid template-fitting in the NSS cells.  

\subsubsection{OL Accuracy: LP \ / \ LR \ / \ IIM-P \ / \ IIM-R / \ IIM-OP / \ IIM-OR}
In Table \ref{tab:ontologyMatrix} we can observe that two out of three engineers created concepts from a pragmatic perspective, thereby neglecting many other ontologically valid, yet not useful, concepts. This is quite expected also when domain experts are involved in engineering ontologies. When we observe the topology generated by DLOL, we find that it does not take a pragmatic approach (since it has no domain-specific rules). Hence, it identified a total number of concepts close to what engineer A did (which is significantly more than the other two engineers). In this sense, DLOL is more exploratory in nature. On the other hand, if we look at the performance of FRED we see that it has failed to identify many primitive concepts. Interestingly, because of its reliance on DRS (and other linguistic frames), it generates a lot of non IS-A relations that ideally should have been recognized as IS-A type relations. This increases the number of redundant/incorrect derived concepts (a potential threat to reasoning efficiency). When we analyzed Text2Onto we observed that it fails to produce any derived concepts. This is because, as a framework, Text2Onto heavily utilizes the IS-A relation-establishing algorithm, while rejecting several possible non IS-A relations that are equally important to identify in terms of knowledge discovery. Also, both Text2Onto and FRED were unable to generate definitions that involve union/intersection operators. For example, in the sentence: ``\textit{John is intelligent}", we need a derived concept labeled \textit{IntelligentPerson} which is an intersection of the concept \textit{Person} and \textit{IntelligentThing}. This cannot be generated by both FRED and Text2Onto. Finally, while analyzing LExO we found out that it generates a lot more number of derived concepts than primitive concepts. This suggests that much of the generated ontology constitutes definitions (rather than axioms). However, we also found that most of the definitions have been incorrectly learned mostly due to: (a) interpretation of \textit{is a} lexicons as equivalence (i.e. \textit{same as}), (b) lack of lexical normalization of \textit{is a} type lexical variations, and (c) insensitivity to quantifier lexemes.

We then evaluated the accuracy of all the OL tools quantitatively on the WCL dataset using the measures discussed prior (see Table \ref{OLAccuracy}). In terms of term extraction capacity, DLOL achieved an \textit{average} score of 0.726 (in contrast to the next best average score of 0.6003 by FRED). In terms of structural accuracy, DLOL achieved a significant improvement over the other tools with an average IIM-R of 0.5056 (in contrast to the next best average score of 0.2736 by FRED). In terms of the degree of true positives (i.e. noise) associated with learned ontologies, Text2Onto performs best, both in terms of LP and IIM-P, with an average score of 0.7357 and 0.5196 respectively. In comparison, DLOL achieved average precision scores of 0.5633 (LP) and 0.3240 (IIM-P). The high performance of Text2Onto can be attributed to the fact that: (i) it is not an exploratory tool (i.e no derived concepts), and (ii) it learns the top-level concepts where the diversification of sub-concepts is least. Hence, noise due to additional learned concepts is low in Text2Onto resulting in high precision but low recall.

\begin{table}[t]
	\centering
	\caption{Degree of Semantic Preservation: Comparative Analysis on Community IS-A Dataset}
	
	\smallskip 
	\begin{tabular}{|c|c|c|c|c|}
		\hline
		\textbf{No} & \textbf{LExO} & \textbf{FRED} & \textbf{Text2Onto} & \textbf{DLOL} \\
		\hline
		Trivial & 14/15 & 9/15 & 13/15 & 0/15 \\
		\hline
		Non-Trivial & 56/56 & 40/56 & 56/56 & 0/56 \\
		\hline
		\textbf{Partial} & \textbf{LExO} & \textbf{FRED} & \textbf{Text2Onto} & \textbf{DLOL}  \\
		\hline
		Trivial & 1/15 & 0/15 & 0/15 & 1/15 \\
		\hline
		Non-Trivial & 0/56 & 12/56 & 0/56 & 2/56 \\
		\hline
		\textbf{Yes} & \textbf{LExO} & \textbf{FRED} & \textbf{Text2Onto} & \textbf{DLOL}  \\
		\hline
		Trivial & 0/15 & 6/15 & 2/15 & 14/15 \\
		\hline
		Non-Trivial & 0/56 & 4/56 & 0/56 & 54/56 \\
		\hline
	\end{tabular} %
	\label{tab:ComparativeAnalysisCustomerDataset}%
\end{table}%

\subsubsection{Degree of Semantic Preservation}
To analyze the degree of semantic preservation of all the compared OL tools, we tested them on the community dataset. We observed that, out of 15 identified IS-A types, the most number of correct semantic preservation (apart from DLOL, which happens to be the best by design) is achieved by FRED (40\% accuracy) followed by Text2Onto (13\% accuracy) (see Table \ref{tab:ComparativeAnalysisCustomerDataset}). When we evaluated the performance on non-trivial IS-A type sentences we observed that FRED again performed best (after DLOL) with 7\% accuracy (tested on 56 distinct cases) in contrast to 96\% accuracy by DLOL. It has to be noted over here that none of the compared OL tools where designed with an extensive study of IS-A type sentences. 

\begin{figure}
\begin{minipage}[t]{0.49\linewidth}
\centering
\includegraphics[height=2in, width=2.4in]{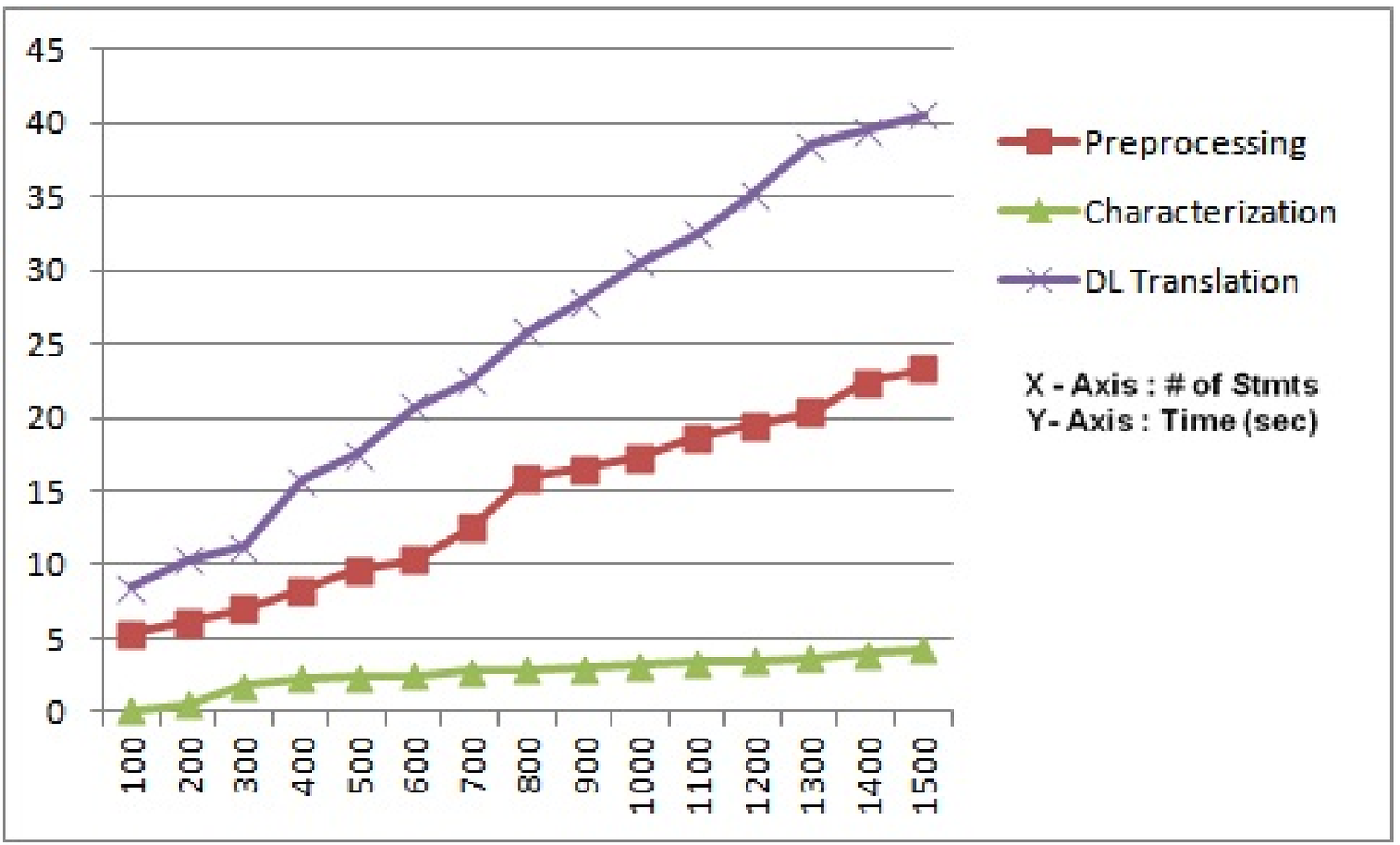}
\caption{$DLOL_{IS-A}$ Runtime Performance}
\label{fig:runtimeAccuracy}
\end{minipage}%
\hfill \hfill
\begin{minipage}[t]{0.49\linewidth}
\centering
\includegraphics[height=2in, width=2.4in]{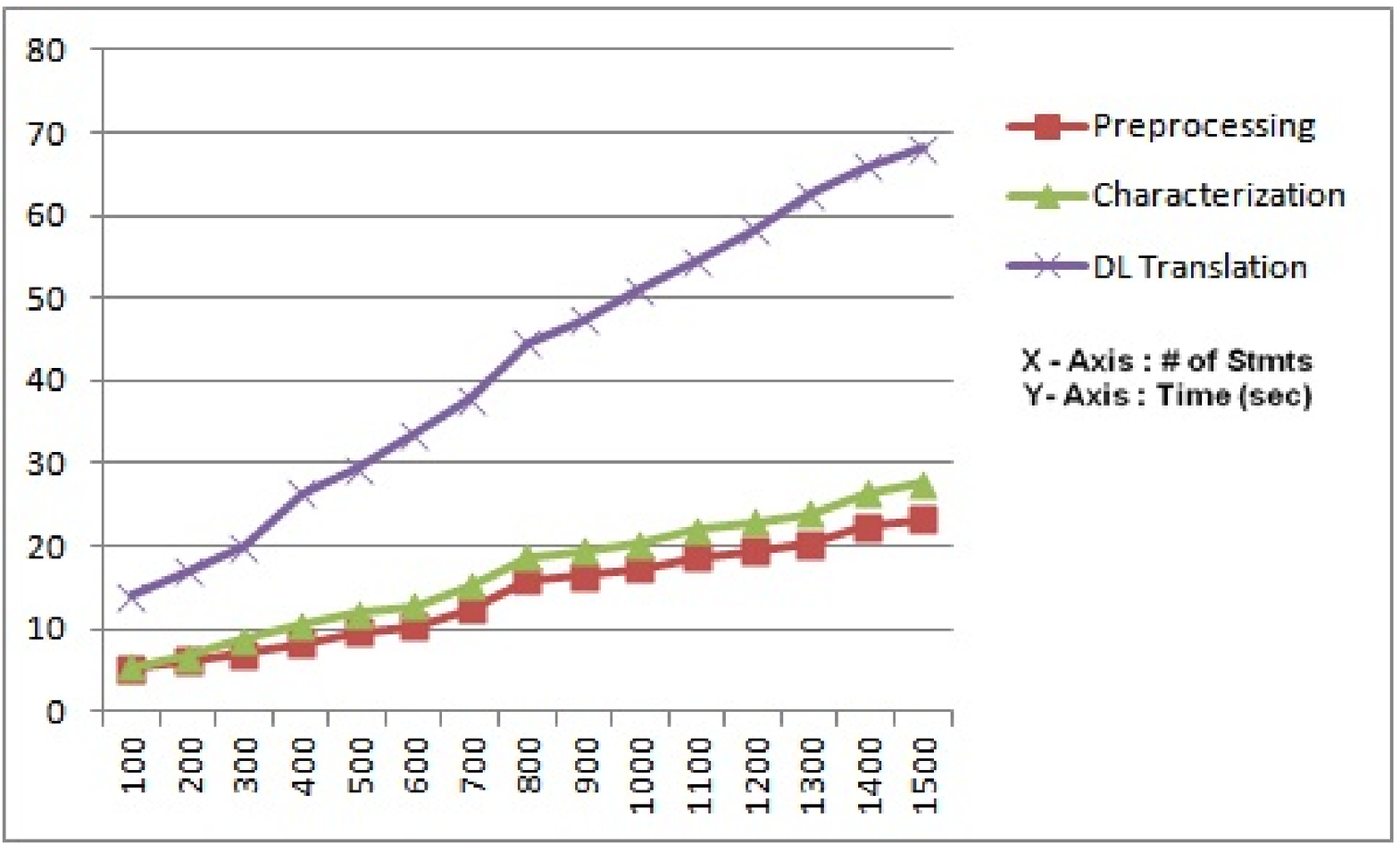}
\caption{$DLOL_{IS-A}$ Cumulative Runtime Performance}
\label{fig:cumulativeRuntimeAccuracy}
\end{minipage}
\end{figure}

\subsubsection{Runtime Efficiency}
We analyzed the runtime efficiency of DLOL for the preprocessing phase (triple extraction, singularization, \& normalization), the characterization phase, and the DL translation phase (i.e. generation of the OWL file) individually. For the experiment we once again chose the WCL dataset. We divided the corpus into 15 sets. We saw that without preprocessing and characterization DLOL takes 8.50 seconds for a set of 100 sentences and 40.62 seconds for a set of 1537 sentences (Figure \ref{fig:runtimeAccuracy}). DL translation, on an average, takes 8.29 seconds more than that of preprocessing phase and characterization phase together (Figure \ref{fig:runtimeAccuracy}). Cumulative runtime of the entire process is given in Figure \ref{fig:cumulativeRuntimeAccuracy}. The total reasoning time by FACT++ (for further axiom induction and consistency checking) after the OWL file generation was 0.239 ms. It is to be understood that web-scale knowledge base generation takes place in a distributed environment with high-end machines and hence, the figures only represent the approximate growth rate of DL translation and also how the translation phase compares to the other two phases. Also, it is to be noted that the entire process is off-line in nature.

\section{RELATED WORK}
\label{relatedWork}
There has been significant literature over the last decade on the problem of Ontology Learning (OL). Most of these works can be categorized into two approaches as discussed earlier: (i) \textit{Light-weight Ontology Learning}, and (ii) \textit{Formal Ontology Learning}. Light-weight OL from text documents is arguably the most widely used approach in the field of OL \cite{4wong2012ontology}. It can be further divided into three general approaches: (i) \textit{distributional semantics based}, (ii) \textit{lexico-syntactic pattern based}, and (iii) \textit{deep linguistic analysis based}. It is to be noted that the general disadvantage of light-weight OL of not being able to generate definitional T-Box (and corresponding A-Box), as discussed in the introduction section, is inherent in all the three approaches. This is where $DLOL$ parts away significantly from light-weight OL approaches. We first discuss light-weight OL in the following three sub-sections for providing a contrasting perspective, and then conclude the section with a discussion on formal OL. 

\subsection{Light-weight OL: Distributional Semantics}
\label{lightOL}
Research on light-weight OL heavily draws models and methodologies from the area of statistical NLP with the assumption of \textit{distributional semantics}. The central objective is to model a similarity measure for concept comparison, thereby creating semantic spaces derived from two alternative models: (i) vector-space models (VSM), and (ii) probabilistic models. In VSM approaches, concept similarity is computed using distance (or similarity) measures in high-dimensional concept vector space. Such computation can be based on: (i) \textit{angular} (such as cosine similarity \cite{15fortuna2006semi}, Jaccard similarity \cite{16doan2003learning}), (ii) \textit{matrix operation} based (such as term-document based matrix models like LSA \cite{17deerwester1990indexing} or term-context terms based matrix models like HAL \cite{18lund1996producing}). Similar concepts are then clustered into taxonomies using variations of known hierarchical clustering algorithms. An early work on OL as an end-to-end application was proposed in \cite{35lin2002concept,36caraballo1999automatic} where noun terms were organized into hypernymic trees using bottom-up hierarchical clustering algorithm. In \cite{37pantel2002document} a clustering based algorithm called CBC was proposed for generating concept lists of similar instances with sense discrimination. However, such lists were unlabeled and hence, not useful in applications such as question-answering as observed in \cite{38pantel2004automatically}.  

While many of the distributional semantics based approaches neglect the word order for generating the semantic space, some advanced works incorporate the order in their measure \cite{19jones2007representing}. Semantic spaces that are created in these ways are mostly language independent and do not take into account the syntactic structure within and across sentences. This leads to poor semantic similarity computation \cite{20fodor1995comprehending}. Other advanced techniques extend vector-space models with morpho-syntactic information such as POS tagging as in \cite{21widdows2003unsupervised}, shallow syntactic analysis as in \cite{22curran2002scaling}, and complete linguistic parsing as in \cite{23lin1998automatic}. The key idea is to select context words that have meaningful syntactic relationship with the target words (i.e. concepts whose similarity has to be computed). A general algorithmic framework for computing such extended semantic space is proposed in \cite{6pado2007dependency}. In \cite{38pantel2004automatically} a top-down clustering based approach was proposed where semantic classes (generated using the CBC algorithm) are automatically labeled using syntactic pattern based class signatures. This resolved the labeling issue to some extent. However, the algorithm uses a variation of K-means and hence, can only generate 1-level taxonomy. An ant-based clustering approach can be found in \cite{39wong2007tree} where concepts are identified as part of the OL process using Normalized Google Distance (NGD) \cite{40cilibrasi2007google} and a new $n^{\circ}$ of Wikipedia similarity measures. However, this work cannot be categorized as text corpus based OL but rather, concept hierarchy induction on collection of terms. Another more recent interesting approach can be found in \cite{42fountain2012taxonomy} where Hierarchical Random Graph (HRG) based lexical taxonomy learning has been proposed. In this work the input is a text corpus and a set of given terms over which a semantic network is generated using distributional semantic similarity measure. An HRG is then fit over the generated semantic network using MLE and Monte Carlo Sampling. The claim is that the technique outperforms popular flat and hierarchical clustering algorithms. Other probabilistic and information theoretic models have also been proposed as in \cite{43tang2009towards} where OL is done on top of folksonomies as input. Alternative non-clustering based techniques have been proposed that fall under the general category of distributional semantics based OL. One work has used Formal Concept Analysis (FCA) for concept identification and hierarchical concept group formation \cite{41cimiano2005learning}.

\subsection{Light-weight OL: Lexico-Syntactic Pattern}
\label{LightOLLexico}
Apart from bag-of-word and syntactic pattern based techniques described in the previous section, there exists an alternative set of approaches that is based on lexico-syntactic pattern mining, where external lexical resources (such as WordNet, external lexical ontologies/taxonomies, semantic labeling rules, etc.) are extensively used for determining semantic similarity between concepts. This particular research direction can be sub-classified into six broad categories: (i) supervised learning based, (ii) unsupervised learning based, (iii) semi-supervised learning based, (iv) graph based, (v) probabilistic linguistic structure instantiation based, and (vi) rule based. A general problem with all such approaches is that the semantics of many linguistic nuances (see section \ref{is-a type} for a discussion on such cases in the context of IS-A type sentences), that are innate in natural language, cannot be represented adequately. Also, there are limitations in terms of semantic translation of quantifiers, modifiers, and lexical variations of the widely accepted linguistic patterns. It is to be noted that both these limitations have been identified and treated in this paper. 

A full-fledged lexico-syntactic OL pipeline would require to solve certain NLP tasks such as: (i) relation extraction \cite{28moldovan2012polaris, 29girju2006automatic, 30pantel2006espresso, 31kozareva2008semantic, 32etzioni2011open}, (ii) named entity recognition \cite{33etzioni2005unsupervised}, and (iii) term extraction \cite{34sclano2007termextractor}. An important work in supervised learning based OL is found in \cite{44snow2006semantic} where supervised classifiers were trained on hyponymy and cousin relation rich dataset and then tested on unknown sentences that had to be classified into a particular relation type and fit onto the WordNet semantic network. Another supervised learning based ontology learner, called JAGUAR, is proposed in \cite{45balakrishna2010semi} where six primary syntactic patterns where semantically labelled together with five verb patterns. Classifiers were then trained on text containing such patterns so as to classify sentences into one of the 26 semantic relations discussed in \cite{28moldovan2012polaris}. \cite{balakrishna2013automatic} extend the mentioned framework \cite{45balakrishna2010semi,28moldovan2012polaris} with knowledge classification algorithm to classify concept in the existing hierarchy. A lexico-syntactic based unsupervised ontology learner has been described in \cite{33etzioni2005unsupervised}. This learner is basically the sub-class extractor component of the KnowItAll system \cite{46etzioni2004web}. This sub-class extractor uses the eight IS-A Hearst pattern rules described in \cite{47hearst1992automatic} together with web-search match results of linguistic patterns to generate new candidate terms for sub-class identification. \cite{mukherjee2014domain} describe an unsupervised framework called ``Domain Cartridge'' to construct shallow domain ontology. Framework provide the mechanism to identify concepts and categorize the relation among concepts into four categories. A semi-supervised taxonomic induction framework is proposed in \cite{48yang2008human}. In this work an ontology metric has been formulated that includes term context, co-occurrence, and lexico-syntactic patterns. This metric is then used to induce a lexicalized taxonomy given a set of concepts. In \cite{huang2014omit}, a semi-automated ontology development for medical domain has been proposed. In this work, an ``backbone'' ontology is created using domain expertise with existing bio-ontologies which is then aligned with another different knowledge base to find equivalent concept among ontologies. Properties of equivalent concept are augmented in ``backbone'' ontology. Perhaps one of the most popular benchmark OL framework is Text2Onto \cite{52cimiano2005text2onto}. Text2Onto is based on instantiation of probabilistic linguistic structures called POM (Probabilistic Ontology Model). POM structures are language independent and can be easily transported to RDFS/OWL/F-Logic by ontology engineers. POMs are essentially seven types of linguistic frames (or patterns) as defined in Gruber's frame ontology \cite{53gruber1993translation}. They are probabilistic in the sense that an instance membership in a particular POM is associated with a degree of belief based on evidences in the corpus. Linguistic analysis of text corpus is based on the JAPE transducer \cite{54maynard2001named} for matching Hearst patterns and POM patterns in order to instantiate. A distinguishing feature of Text2Onto is semi-automatic revisioning of ontology as the ontology evolves learning new instances. A detailed experimental comparison with our proposed tool, $DLOL$, is given in Tables \ref{tab:ontologyMatrix}, \ref{OLAccuracy}, and \ref{tab:ComparativeAnalysisCustomerDataset}. 

A graph based approach can be found in \cite{49kozareva2010semi}. Over here the proposed system assumes that a set of root primitive concepts will be given to the system as input. It then extracts new term pairs having the root terms as one of the pair element from the Web such that the sentences extracted contain Hearst patterns. This set of term pairs is then searched on the Web again to get all triples containing the pairs. The hyponymy graph is then generated by selecting terms that have a minimum IS-A out-degree with respect to a root term. The limitation is that root terms has to be known apriori which works only in domain specific applications. Another very recent graph based ontology learner, called OntoLearn Reloaded, has been proposed in \cite{50velardi1ontolearn}. OntoLearn works in five steps. First it extracts terms from a document corpus using TermExtractor tool \cite{34sclano2007termextractor}, after which a set of definitional sentences containing the extracted terms is selected using the WCL classifier \cite{51navigli2010learning}. These definitional sentences are then filtered according to domain resulting in a noisy hypernymic graph. A pruning procedure is carried on the noisy graph to generate a tree-like taxonomy. As a final step a DAG based taxonomy is formed by reviving incorrectly deleted edges during the pruning process. Graph based ontology learning like TaxoFinder \cite{kang2016taxofinder} learns taxonomy from the given text corpus. TaxoFinder is divided into three parts (1) concept extraction (2) association among the concepts to form a graph and (3) taxonomy generation from the graph. Concept are extracted using CFinder \cite{kang2014cfinder} which make use of linguistic patterns, domain-specific knowledge and statistical techniques to identify and weight key concept specific to the domain. Extracted concept are connected to each other depending on the strength of the association. Association strength is determined using a context of window sentences - number of sentences above and after the given pair of concept and similarity of the words. Graph obtained in this manner is processed to derive taxonomy such that it maximizes the overall association strength among all concepts. Finally, several OL techniques have been proposed that it purely rule based with application of linguistic heuristic and supporting knowledge sources such as Wikipedia \cite{55suchanek2008yago, 56ponzetto2009large, 57ponzetto2011taxonomy}.

\subsection{Light-weight OL: Deep Linguistic Analysis}
\label{LightOLLinguistic}
A third approach for light-weight OL is deep linguistic analysis based. It heavily draws techniques prevalent in \textit{computational semantics}. Computational semantics largely involves the algorithmic feasibility and mechanism for translating NL sentences into semantically equivalent formal language expressions so as to represent its meaning. Several formal semantic theories have been proposed in this direction. Perhaps the most influential pioneering theory is what is known as \textit{Montague Grammar} \cite{72montague1970english}. Montague Grammar is a model theoretic representation of NL and its computational feasibility is proven by the applicability of $\lambda$-calculus \cite{73church1932set} to construct the semantics over a parse tree generated by the underlying grammar. Another alternative computational technique for semantic construction is \textit{feature structures} \cite{74kasper1986logical}. More recent semantic theories, mostly proposed to deal with semantic interpretation across sentence boundaries, include Discourse Representation Theory \cite{75kamp1981theory} and File Change Semantics \cite{76heim1983file}. A computational tool for DRT based semantic representation, called BOXER, has been proposed in \cite{77bos2008wide}. This tool generates semantic representations in DRS form (Discourse Representation Structure) after syntactically parsing sentences using CCG (Combinatory Categorial Grammar). However, the first and perhaps the only OL tool to incorporate BOXER, and for that matter any such linguistic analysis tool, was FRED \cite{13presutti2012knowledge,Gangemi2016}. FRED also incorporates linguistic frames and ontology design patterns with Wikipedia as an external supporting source. However, it does not attempt to generate any expressive ontology, but rather an RDF graph structure. This limits its capability to support complex and sophisticated reasoning for knowledge discovery. We also observed that it cannot adequately represent the semantics of many linguistic nuances, as discussed in section \ref{is-a type}. A thorough investigation of FRED has been given in Tables \ref{tab:ontologyMatrix}, \ref{OLAccuracy}, and \ref{tab:ComparativeAnalysisCustomerDataset}. 
\subsection{Formal OL}
\label{formalOL}
There is considerable shortage of literature that can be truly classified as formal OL (as has been observed in \cite{4wong2012ontology}). Early works related to formal OL primarily focused on inducing disjoint axioms in a given ontology (learned or manual) so as to debug modeling errors. A heuristic approach (called \textit{semantic clarification}) based on the assumption of \textit{strong disjointness} \cite{58cornet2003using} has been proposed in \cite{59schlobach2007debugging}. There has also been several machine learning (ML) based approaches to disjoint axiom induction. One supervised learning based framework called LEDA was proposed in \cite{60meilicke2008learning, 61fleischhacker2011inductive} where manually created disjoint axioms were used as the training dataset. Another ML based approach can be found in \cite{62haase2008ontology} where unsupervised mining of disjoint axioms from text corpus was proposed. Other alternative techniques for disjoint axiom induction include one work based on FCA \cite{63baader2007completing} and another, called DL-Learner, based on inductive logic programming \cite{64lehmann2009dl}. There has been a recent work on association mining based disjoint axiom induction \cite{65volker2011statistical}. This work was further extended in \cite{66fleischhacker2011inductive} where three alternative approaches have been proposed - two based on negative association mining (i.e. confidence measure of whether membership in one class implies non-membership in another class) and one based on statistical correlation analysis (using $\phi$-coefficient to measure strength of disjointness between two concepts where complete disjointness has a $\phi$ value of -1.0). Apart from disjoint axiom induction there has been one work that focused on negation axiom induction \cite{67lehmann2011class} where an ontology repairing tool called ORE was proposed.  However, none of the works described so far can be described as end-to-end formal OL on text corpus but all are, rather, formal ontology enrichment/correction methodologies.

One of the earliest approaches towards formal OL can be found in \cite{68davis2009deep}. In this work a very novel approach for probabilistically inducing formal structures (First Order Predicate Logic (FOL) derivative) from text has been outlined. The technique is based on the principle of \textit{deep transfer learning} which postulates that a model can generalize structures from one domain corpus (called \textit{source}) and be plugged into another model that generalizes in a completely different domain corpus (called \textit{target}) to increase both the efficiency and accuracy of target model. For deep transfer, Markov Logic Network induction from natural text was used, where FOL formulae are represented as graph structures and clique structures within this graph (representing the set of such FOL formulae) are modeled as Markov network. This work was extended further in \cite{69davis2011deep}. Another similar work can be found in \cite{70poon2010unsupervised}. Here a tool called OntoUSP was proposed where the input to the system was a dependency-parsed text and the output is a probabilistic auto-learned ontology with IS-A and part-of relations. The semantic parsing required for inducing high order Markov formulae \cite{71domingos2009markov} is unsupervised. However, all such techniques do not use significant research outcomes that have been reported in theoretical and computational semantics within the linguistic community and also exclude important results in the area of NL understanding based question-answering that are closely related to the problem of formal OL.

\section{Discussion \& Future Work}
Although most of the identified cases of IS-A type sentences can be processed by DLOL yet there are certain exceptional categories that we would like to highlight in this section. One very interesting case can be understood from the following two sentences that are syntactically equivalent: (i) ``\textit{A cup of coffee will be wonderful.}", and (ii) ``\textit{The son of God is great}". In the first sentence the primary subject is \textit{coffee} while in the second sentence the primary subject is \textit{son}. Another case of ambiguity may arise from the contextual usage of the indeterminate article before a noun. For example, in the sentence ``\textit{A planet is round}" the article denotes the general class of \textit{Planet} while in the sentence ``\textit{A boy is tall}" the article denotes a specific instance of the class \textit{Tall Boy}. We leave the disambiguation of these cases as a future work. Apart from further work on IS-A type sentences, an extensive and thorough work on non-ISA type sentences is ongoing. We have observed that more detailed parsing is required for extending the sentence simplifier, new base rules are required for DL translation, and new spatio-temporal rules have to be incorporated for modal reasoning. All of these ongoing effort is directed towards the completion of a working prototype, called \textit{DLOL(F)} (where \textit{F} stands for `\textit{full}").

\section{Conclusion}\label{sec:Conclusion}
In this paper we propose a Description Logics (DL) based formal ontology learning (OL) tool, called \dlol, that works on factual non-negative IS-A sentences. We argue that OL on IS-A type sentences is non-trivial. In support to this argument, we evaluated the accuracy of \dlol \ on IS-A datasets, against three state-of-the-art tools, and observed that it had an average LR improvement of 21\% (approx.) and an average IIM-Recall improvement of 46\% (approx.) with respect to next best performing tool.

\section{Acknowledgement}
We would like to thank Prof. Thomas Lukasiewicz from University of Oxford for valuable insights and suggestions.

\bibliographystyle{elsarticle-num}
\bibliography{aij_dlol_ref}

\newpage
\end{document}